\documentclass[10pt,twocolumn,letterpaper]{article}

\usepackage{iccv}
\usepackage{times}
\usepackage{epsfig}
\usepackage{graphicx}
\usepackage{amsmath}
\usepackage{amssymb}

\usepackage{caption}
\usepackage{subcaption}
\usepackage{multirow}
\usepackage{here}

\makeatletter
\def\Hline{
  \noalign{\ifnum0=`}\fi\hrule \@height 2.\arrayrulewidth \futurelet
  \reserved@a\@xhline}
\makeatother

\iccvfinalcopy 

\def\iccvPaperID{****} 

\begin{document}

\title{What Do Adversarially Robust Models Look At?}

\author{
    Takahiro Itazuri\thanks{First two authors contributed equally.}${}^{*1}$,
    Yoshihiro Fukuhara${}^{*1}$, 
    Hirokatsu Kataoka${}^{2}$, 
    Shigeo Morishima${}^{3}$ \vspace{1mm}\\
    ${}^{1}$Waseda University, 
    ${}^{2}$National Institute of Advanced Industrial Science and Technology,\\ 
    ${}^{3}$Waseda Research Institute for Science and Engineering
}

\maketitle

\begin{abstract}
In this paper, we address the open question: ``What do adversarially robust models look at?''
Recently, it has been reported in many works that there exists the trade-off between standard accuracy and adversarial robustness.
According to prior works, this trade-off is rooted in the fact that adversarially robust and standard accurate models might depend on very different sets of features.
However, it has not been well studied what kind of difference actually exists.
In this paper, we analyze this difference through various experiments visually and quantitatively.
Experimental results show that adversarially robust models look at things at a larger scale than standard models and pay less attention to fine textures.
Furthermore, although it has been claimed that adversarially robust features are not compatible with standard accuracy, there is even a positive effect by using them as pre-trained models particularly in low resolution datasets.
\end{abstract}

\section{Introduction}
Deep neural networks (DNNs) have dramatically improved the accuracy of various computer vision tasks.
Especially in image classification, the top-5 error rate of ResNet~\cite{he2016deep} reached 3.57\% and already exceeded human performance (5.1\%)~\cite{russakovsky2015imagenet} on ImageNet~\cite{imagenet}.
On the other hand, it is well known that even state-of-the-art DNNs~\cite{he2016deep, zagoruyko2016wide, szegedy2016rethinking, yu2017dilated} are highly vulnerable to imperceptible and well-sought perturbations, called {\it adversarial examples}~\cite{szegedy2013intriguing}.
The existence of adversarial examples is an obstacle for the real use of DNNs from the aspect of security, and the analysis of this phenomenon is imperative. 

\begin{figure}
    \centering
    \begin{subfigure}[b]{0.5\textwidth}
        \centering
        \includegraphics[width=\textwidth]{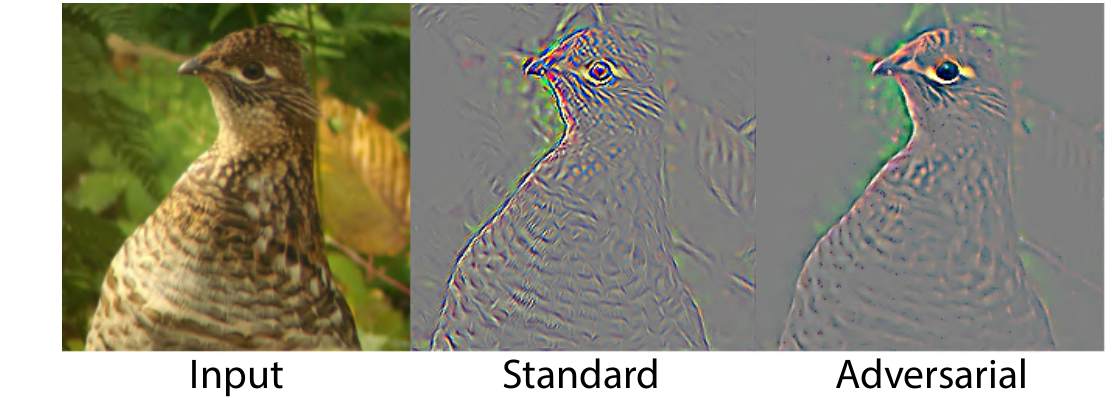}
        \caption{\normalsize Sensitivity map visualization}
    \end{subfigure}
    
    \vspace{2ex}%
    
    \begin{subfigure}[b]{0.5\textwidth}
        \centering
        \includegraphics[width=\textwidth]{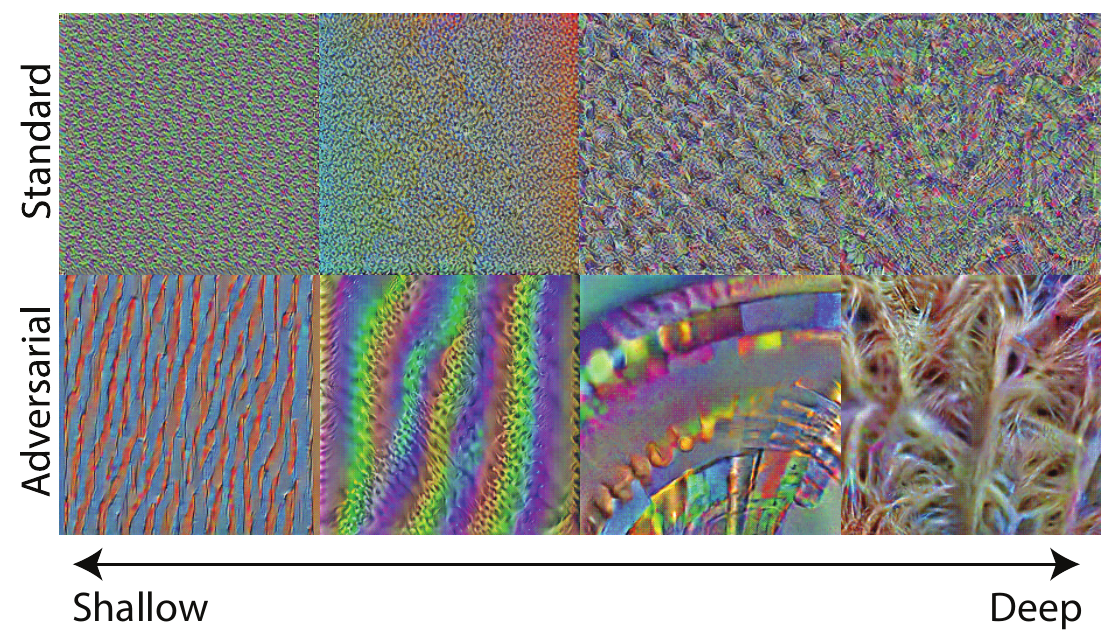}
        \caption{\normalsize Layer visualization}
    \end{subfigure}
    \caption{
         We clarify the difference between the representation obtained by adversarially trained and standard trained models through various experiments.
        (a) Sensitivity map visualization of a standard trained (middle) and an adversarially trained (right) ResNet-50 by guided backpropagation~\cite{springenberg2015striving}. While the both models are sensitive to strong edges, the adversarilly trained one does not react to fine textures.
        (b) Layer visualization \cite{erhan2009visualizing} of the standard trained (top row) and the adversarially trained (bottom row) ResNet-50. Each column represents the same depth visualization.
        Visualization results suggest that the adversarially trained model tries to capture larger scale patterns than the standard trained model.
        \vspace{-3mm}
        %
    }
    \label{fig:teaser}
\end{figure}
Due to these strong demand, many works have addressed it from various points of view~\cite{szegedy2013intriguing, goodfellow2015explaining, tanay2016boundary, samangouei2018defensegan, gilmer2018adversarial}.
As a result of many efforts, many interesting findings of DNNs have been accumulated in this area.
One of them is that training models to be robust to adversarial perturbations leads to a reduction of standard accuracy~\cite{kurakin2016adversarial, kurakin2017adversarialmachine, madry2018towards, su2018robustness, tsipras2019robustness}.
This phenomenon is contrary to the simple expectation~\cite{torkamani2014robustness, goodfellow2015explaining, miyato2018virtual} that training on adversarial examples as ultimate form of data augmentation can improve generalization performance in the standard classification.
Tsipras et al.~\cite{tsipras2019robustness} suggested that this trade-off is caused by the fact that adversarially trained models obtain fundamentally different feature representations than standard trained models.
Therefore, analyzing the features acquired by the adversarially trained models and clarifying what features are used for their decision making are useful not only for designing stronger defense methods but also for a better understanding of the phenomenon of adversarial perturbations.
However, the nature of the feature representations acquired by adversarially trained models has not been sufficiently elucidated yet.
In this work, we reveal the property of the adversarially robust features through comprehensive experiments and visualizations (Figure~\ref{fig:teaser}).

our contributions are summarized as follows:
\begin{itemize}
    \item We reveal that adversarially robust features look at things at a larger scale than standard models and pay less attention to fine textures through various visualization methods and quantitative evaluations where we expose them to various artificial transformations.
    \item We conduct a comprehensive study on various datasets to reveal how adversarially robust features are beneficial for improving standard accuracy and show that they are useful as pre-trained models and especially have a positive effect on accuracy in low resolution datasets.
\end{itemize}

To promote reproducible research, we will release our code
including reimplementation of multiple adversarial attacks and tools for visualizing CNN.

\section{Related Work}
\subsection{Trade-off Between Robustness and Accuracy}
So far, various defense methods~\cite{goodfellow2015explaining,papernot2016distillation,cisse2017parseval,metzen2017detecting,miyato2018virtual,tramer2018ensemble,moosavi2019robustness} have been proposed against attack methods~\cite{goodfellow2015explaining,moosavi2016deepfool,papernot2016limitations,carlini2017towards,kurakin2017adversarialexamples,moosavi2017universal,madry2018towards,song2018constructing,athalye2018synthesizing,modas2019sparsefool}.
However, most of them were subsequently shown to be ineffective~\cite{carlini2017adversarial,athalye2018obfuscated,uesato2018adversarial} and adversarial training~\cite{goodfellow2015explaining, madry2018towards}, which is the simplest one, seems to be most effective so far.
Unfortunately, it has been reported in many works that standard accuracy has to be sacrificed to train adversarially robust models~\cite{kurakin2016adversarial,fawzi2018analysis,dvijotham2018dual,wong2018scaling,su2018robustness,madry2018towards,xiao2019training}.
Fawzi et al.~\cite{fawzi2018analysis} proved upper bounds of the adversarial robustness of classifiers and exhibited the trade-off between standard accuracy and adversarial robustness for a specific classifier families on a synthetic task.
Su et al.~\cite{su2018robustness} conducted a comprehensive experiment on 18 different network architectures that are submitted to ImageNet Challenge~\cite{russakovsky2015imagenet}, and empirically showed that the higher accuracy models are, the more vulnerable to adversarial perturbations.
Tsipras et al.~\cite{tsipras2019robustness} exhibited that the feature representations need to achieve high adversarial robustness are different from those need to achieve high standard accuracy, and suggested that this difference is the main cause of the trade-off between adversarial robustness and standard accuracy.

\subsection{Interpretation of Deep Neural Networks}
Explaining classification decisions could potentially shed light on the underlying mechanisms of blackbox systems like DNNs.
Due to this potential, various types of visualization and analysis methods have been proposed~\cite{erhan2009visualizing, zeiler2014visualizing, springenberg2015striving, bach2015pixel, sundararajan2017axiomatic, smilkov2017smoothgrad, shrikumar2017learning, selvaraju2017grad}.

The simplest idea is visualizing each layer of trained CNNs directly.
While the first layer can be visualized straightforwardly, it is hard to visualize deeper layers because of the large number of their dimensions.
Erhan et al.~\cite{erhan2009visualizing} calculated the input which maximizes the activation of the layer of interest by gradient ascent.
Zeiler et al.~\cite{zeiler2014visualizing} projected the feature activations back to the input space by using deconvolutional network~\cite{zeiler2011adaptive} which consists of unpooling and deconvolution.

Instead of visualizing layers directly, there is an approach to create a map that reasons where in the input image contribute to the classifier's decision.
Simonyan et al.~\cite{simonyan2013deep} proposed ``sensitivity map'' by propagating the output of the target class backward, but this simple backpropagation leads to noisy results.
To make the sensitivity map clearer, Springenberg et al.~\cite{springenberg2015striving} proposed guided backpropagation technique that impose restrictions to the gradient and Smilkov et al.~\cite{smilkov2017smoothgrad} proposed SmoothGrad that averages results of multiple noise added inputs.
These pixel-space gradient visualization methods are high-resolution and highlight fine-grained details in the image, but are not class-discriminative.
To localize the class-specific discriminative regions in exchange for its resolution, various methods to generate ``class activation map'' have been proposed (e.g., CAM~\cite{zhou2016learning}, Grad-CAM~\cite{selvaraju2017grad} and Grad-CAM++~\cite{chattopadhay2018gradcam}).
The sensitivity map and the class activation map are often used together to compensate for each other drawbacks.

In contrast to the visualization-based analysis, there are some works which try to elucidate the decision process of networks by investigating the reaction to architecture modifications or peculiar inputs.
Brendel et al.~\cite{brendel2018approximating} proposed a variant of the ResNet-50 architecture called BagNet which imitates the traditional bag-of-feature (BoF) classifiers and claimed that there is noｔ much difference between decision making mechanism of modal DNN classifier and that of traditional BoF classifier on the basis of that BagNet achieved comparable accuracy as state-of-the art deep neural networks.
Geirhos et al., \cite{geirhos2018imagenet} generated images whose texture is replaced by other one’s while maintain the composition and outline using style transfer, and reveals that typical CNN classifier’s decisions are biased towards texture information.

\section{Preliminaries}
\subsection{Adversarial Examples}
Although various types of adversarial examples have been proposed as described in the previous section, hereinafter adversarial examples refer to perturbation-based ones unless otherwise noted.
Given a $K$-class classifier $f: \mathbb{R}^d \to \mathbb{R}^K$ and a pair of a data point $\boldsymbol{x} \in \mathbb{R}^d$ and a ground-truth label $t$, the predicted label is obtained by $\hat{k}(\boldsymbol{x})= {\rm arg~max}_{k} f_k (\boldsymbol{x})$, where $f_k(\boldsymbol{x})$ is the $k$-th component of $f(\boldsymbol{x})$ that corresponds to the $k$-th class.
A set of adversarial perturbations is defined as follows:
\begin{equation}
    \left\{ \boldsymbol{\delta} \in \boldsymbol{\Delta} \; \middle| \; \hat{k}(\boldsymbol{x}+\boldsymbol{\delta}) \neq t \right\},
\end{equation}
where $\boldsymbol{\Delta} \subset \mathbb{R}^d$ is a set of perturbations.
Practically, an adversarial perturbation $\boldsymbol{\delta} \in \mathbb{R}^d$ is given by the following optimization problem:
\begin{equation}
    \max_{\boldsymbol{\delta} \in \boldsymbol{\Delta}} L ( \hat{k}(\boldsymbol{x} + \boldsymbol{\delta}), t ),
\end{equation}
where $L$ is a loss function, e.g., the cross-entropy loss.
In this paper, we focus on the case when $\boldsymbol{\Delta}$ is the set of $\ell_p$-bounded perturbations, i.e. $\boldsymbol{\Delta}=\left\{ \boldsymbol{\delta} \in \mathbb{R}^d \, \middle| \, \|\boldsymbol{\delta}\|_p \leq \varepsilon \right\}$.

\subsection{Adversarial Attack Methods}
Here, we introduce two simple first-order methods that we use for adversarial training and robustness evaluation in this paper. 

\noindent
{\bf Fast Gradient Sign Method.}
One of the simplest methods to generate adversarial examples is fast gradient sign method (FGSM)~\cite{goodfellow2015explaining} which is a fast single-step attack by maximizing the loss function in the linear approximation. The perturbation by FGSM is calculated as follows:
\begin{equation}
    \boldsymbol{\delta} = \varepsilon\cdot\mathrm{sign} ( \nabla_{\boldsymbol{x}} L ( \hat{k}(\boldsymbol{x}),y ) ).
\end{equation}

\noindent
{\bf Projected Gradient Descent.}
Projected gradient descent method (PGD)~\cite{kurakin2017adversarialmachine,madry2018towards} is the iterative variant of FGSM.
The perturbation by PGD at a time step $t+1$ is calculated as follows:
\begin{equation}
    \boldsymbol{\delta}^{(t+1)} = \mathcal{P}_{\varepsilon} ( \boldsymbol{\delta}^{(t)} + \alpha \cdot \mathrm{sign} ( \nabla_{\boldsymbol{x}} L ( \hat{k} (\boldsymbol{x} + \boldsymbol{\delta}^{(t)}) ,t ) ) ),
\end{equation}
where $\alpha$ denotes a single step size and $\mathcal{P}_{\varepsilon}(\cdot)$ denotes the projection onto the $\ell_p$-ball of radius $\varepsilon$.

\begin{table*}[t]
    \centering
    \begin{tabular}{c|c|p{6em}|p{6em}|p{6em}|p{6em}} \Hline
        \multirow{2}{*}{Architecture} & \multirow{2}{*}{Setting} & \multirow{2}{*}{\hfil Top-1 Acc [\%] \hfil} & \multirow{2}{*}{\hfil Top-5 Acc [\%] \hfil} & \multicolumn{2}{c}{Robustness Acc [\%]} \\ \cline{5-6}
        & & & & \hfil $\varepsilon=0.005$ \hfil & \hfil $\varepsilon=0.01$ \hfil \\
        \Hline
        \multirow{2}{*}{AlexNet} 
            & STD & \hfil $\mathbf{58.28} \pm \mathbf{0.18}$ \hfil & \hfil $\mathbf{80.60} \pm \mathbf{0.05}$ \hfil & \hfil $2.48 \pm 0.18$ \hfil & \hfil $0.22 \pm 0.07$ \hfil \\
            & ADV & \hfil $50.59 \pm 0.22$ \hfil & \hfil $73.73 \pm 0.07$ \hfil & \hfil $\mathbf{71.55} \pm \mathbf{0.12}$ \hfil & \hfil $\mathbf{46.49} \pm \mathbf{0.04}$ \hfil \\ \hline
        \multirow{2}{*}{ResNet-50}
            & STD & \hfil $\mathbf{75.84} \pm \mathbf{0.31}$ \hfil & \hfil $\mathbf{92.74} \pm \mathbf{0.14}$ \hfil & \hfil $0.78 \pm 0.03$ \hfil & \hfil $0.17 \pm 0.02$ \hfil \\
            & ADV & \hfil $71.52 \pm 0.06$ \hfil & \hfil $90.02 \pm 0.03$ \hfil & \hfil $\mathbf{81.25} \pm \mathbf{0.09}$ \hfil & \hfil $\mathbf{60.75} \pm \mathbf{0.33}$ \hfil \\ \Hline
    \end{tabular}
    \caption{Results of standard accuracy and adversarial robustness on ImageNet in standard training and adversarial training settings. We used $\ell_{\infty}$-PGD attack for robustness evaluation. There exists the trade-off between standard accuracy and robustness accuracy in both AlexNet and ResNet-50.}
    \label{tab:imagenet}
\end{table*}
\begin{figure*}[t]
    \centering
    \includegraphics[width=\textwidth]{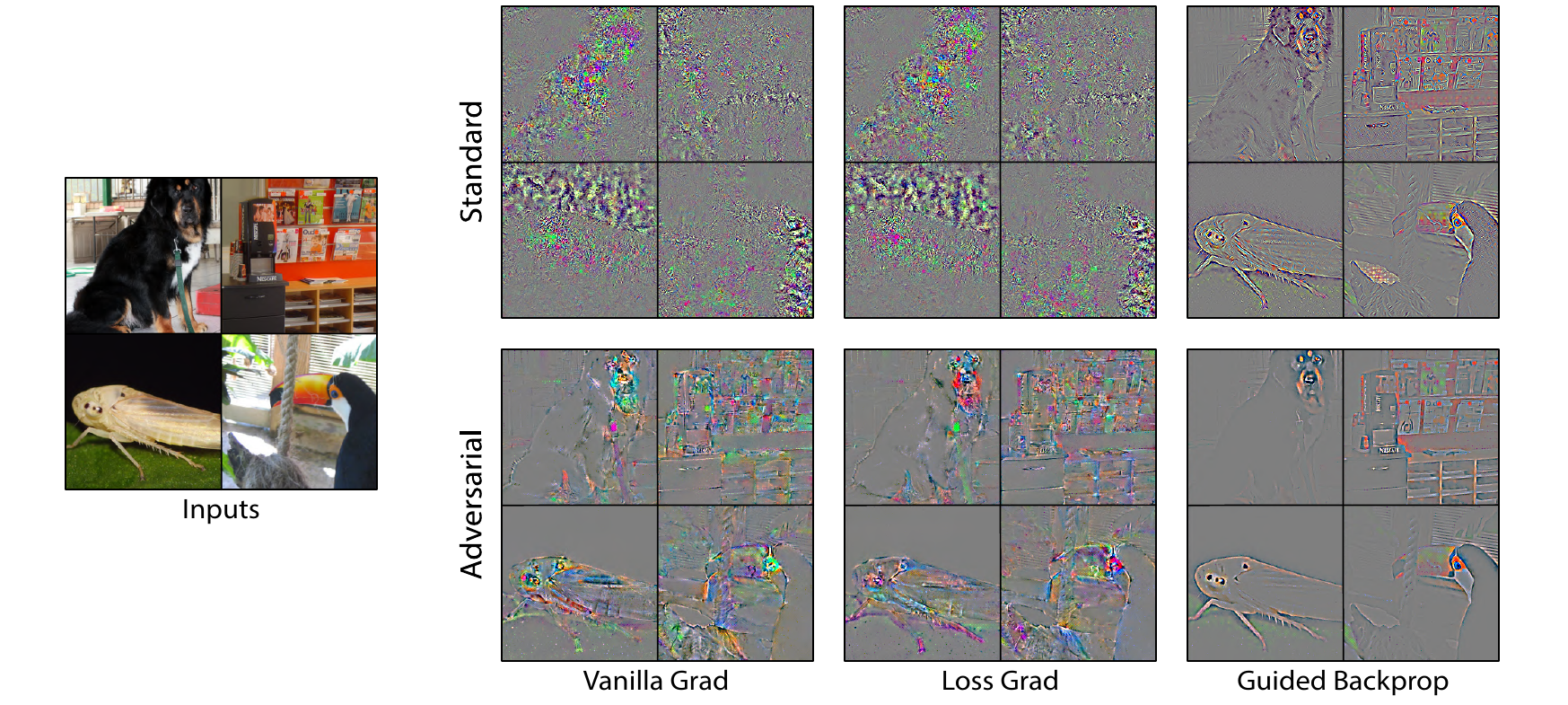}
    \caption{Visualization results of the sensitivity maps of standard trained (top row) and adversarially trained (bottom row) ResNet-50 by Vanilla Grad, Loss Grad and Guided Backprop~\cite{springenberg2015striving}. Additional results are shown in the supplementary materials.}
    \label{fig:map}
\end{figure*}

\subsection{Adversarial Training}
One of the most popular defense approaches is adversarial training~\cite{goodfellow2015explaining, madry2018towards}, which generates adversarial examples to the original training data during the training stage.
Let us consider a classification task with underlying data distribution $\mathcal{D}$.
The goal of standard training is often formulated as finding parameters $\theta$ of the classifier $f$ to minimize the expectation of the loss function:
\begin{equation}
    \min_{\theta} \underset{(\boldsymbol{x},t) \sim \mathcal{D}}{\mathbb{E}} \left[ L ( \hat{k} (\boldsymbol{x}), t ) \right].
\end{equation}
On the contrary, the goal of adversarial training is formulated to minimize the expectation of the adversarial loss:
\begin{equation}
    \min_{\theta} \underset{(\boldsymbol{x},t) \sim \mathcal{D}}{\mathbb{E}} \left[ \max_{\boldsymbol{\delta} \in \Delta} L ( \hat{k} (\boldsymbol{x} + \boldsymbol{\delta}), t ) \right].
\end{equation}
Although adversarial training is effective, this leads to an increase of the training time.
Therefore, more efficient ways of adversarial attacks and adversarial defenses are being sought.

\subsection{Visualization Methods}
In this section, we introduce several visualization methods that we use in our experiments.
All these methods use the gradient information via backpropagation because neural networks are differentiable.
The difference from the training stage is that the optimization is performed on the input with the classifier's parameters fixed.

\noindent
{\bf Vanilla Gradient.}
Vanilla gradient (Vanilla Grad) is the simplest sensitivity map that calculates the gradient of the classifier output for the ground-truth label with respect to the input as follows:
\begin{equation}
    \boldsymbol{S}_{\rm vanilla} (\boldsymbol{x}) = \nabla_{\boldsymbol{x}} f_t (\boldsymbol{x}).
\end{equation}
The sensitivity map by Vanilla Grad tends to be noisy as shown in Figure~\ref{fig:map}.

\noindent
{\bf Loss Gradient.}
Loss gradient (Loss Grad), used in~\cite{tsipras2019robustness}, calculates the gradient of the loss function instead of the classifier output in Vanilla Grad as follows:
\begin{equation}\label{eq:lossgrad}
    \boldsymbol{S}_{\rm loss} (\boldsymbol{x}) = \nabla_{\boldsymbol{x}} L(\hat{k}(\boldsymbol{x}), t).
\end{equation}
Especially in the classification task, the cross entropy loss is often used as the loss function, in which case Equation~\ref{eq:lossgrad} is as follows: 
\begin{align}
    \boldsymbol{S}_{\rm loss} (\boldsymbol{x}) &= - \nabla_{\boldsymbol{x}} \log \left( \frac{\exp f_t (\boldsymbol{x})}{\sum_{k=1}^{K} \exp f_k (\boldsymbol{x})} \right) \nonumber \\
    &= - \nabla_{\boldsymbol{x}} f_t (\boldsymbol{x}) + \sum_{k=1}^{K} y_k (f(\boldsymbol{x})) \nabla_{\boldsymbol{x}} f_k (\boldsymbol{x}),
\end{align}
where $y(\cdot)$ is the softmax function.
The first term is the negative of Vanilla Grad and the second term contains the information other than the ground-truth label.

\noindent
{\bf Guided Backpropagation.}\footnote{Adebayo et al.~\cite{adebayo2018sanity} pointed out that Guided Backprop is misleading because of its independence on the model parameters, but we could not confirm this phenomenon. In detail, we discuss this problem in supplementary materials.}
Compared to Vanilla Grad and Loss Grad, guided backpropagation (Guided Backprop)~\cite{springenberg2015striving} set all the negative gradients to 0 like the ReLU function in order to focus on what kind of feature the neurons detect than what they do not detect.
As a result of performing this operation, the sensitivity map becomes clearer than Vanilla Grad and Loss Grad as shown in Figure~\ref{fig:map}.

\noindent
{\bf Deeper Layer Visualization.}
Erhan et al.~\cite{erhan2009visualizing} proposed a method to synthesize an input that maximizes the activation of a layer of interest without searching for an input pattern from the datasets.
Given a $i$-th filter of a $l$-th layer, the synthesized input $\boldsymbol{x}^{\ast}$ is given by solving the following optimization problem via gradient ascent:
\begin{equation}
    \boldsymbol{x}^{\ast} = \mathop{\rm arg~max}\limits_{\boldsymbol{x} \in \mathbb{R}^d} a^{(l)}_i (\boldsymbol{x}),
\end{equation}
where $a^{(l)}_i$ is sum of the activations of the $i$-th filter of the $l$-th layer.

\section{Experiments}
First, we conducted standard and adversarial training several times from scratch for AlexNet~\cite{krizhevsky2014one} (with batch normalization~\cite{ioffe2015batch}) and ResNet-50~\cite{he2016deep} on ImageNet.
Then we used the momentum stochastic gradient descent (SGD) method for 90 epochs and set the momentum value to 0.9, the batch size to 256, and the initial learning rate to 0.1.
The learning rate is multiplied by 0.1 for each 30 epochs.
In adversarial training, we used $\ell_{\infty}$-FGSM attack ($\epsilon = 0.005$).

Hereafter, standard training and adversarial training are abbreviated as STD and ADV respectively in tables and figures.
Table~\ref{tab:imagenet} shows results of standard accuracy and robustness accuracy (defense success rate against adversarial attacks) on ImageNet in standard training and adversarial training settings.
We can confirm the trade-off between standard accuracy and adversarial robustness, similar to prior works~\cite{su2018robustness,tsipras2019robustness}.

\begin{figure*}[h]
    \centering
    \includegraphics[width=0.95\textwidth]{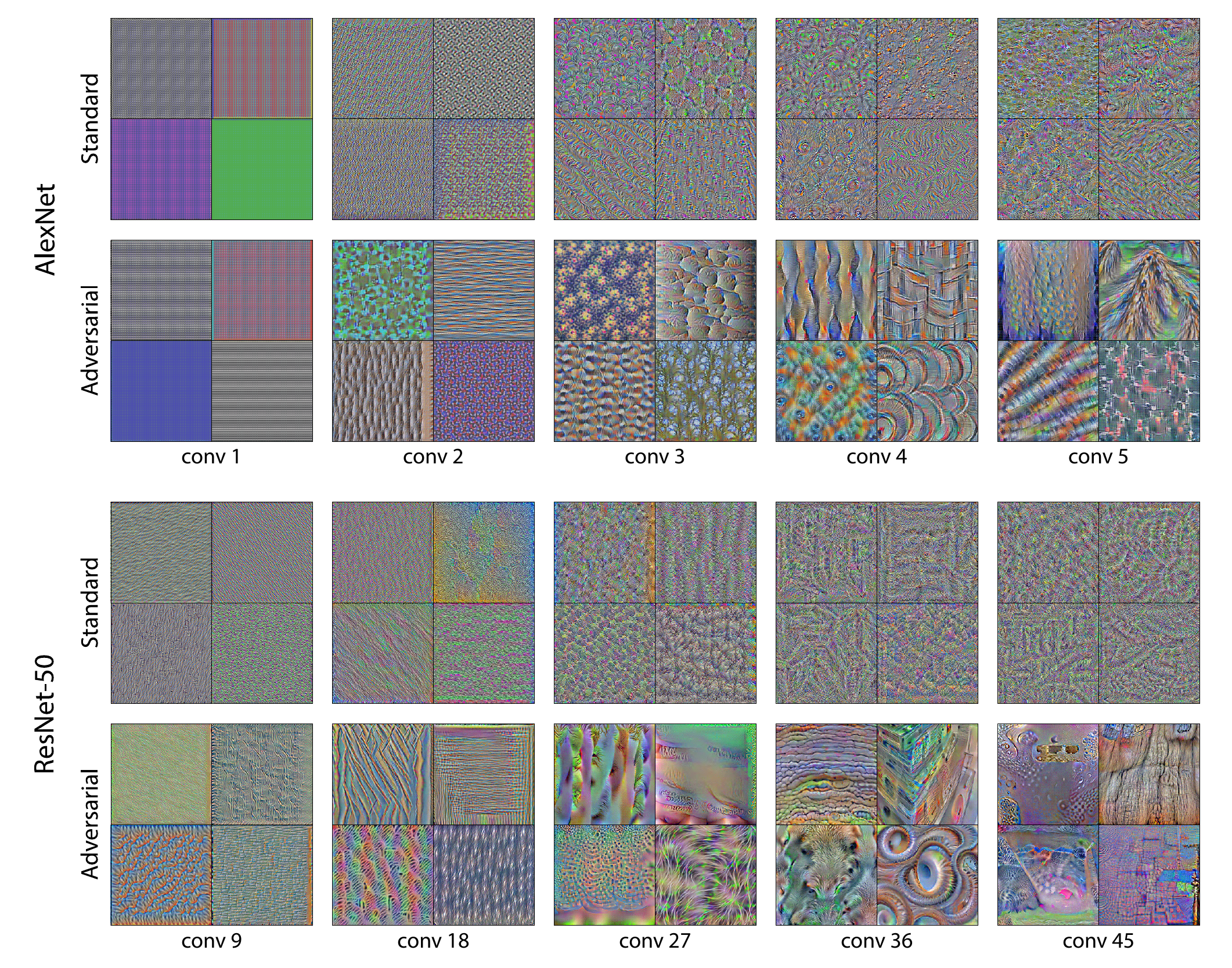}
    \caption{Deeper layer visualization of standard trained and adversarially trained models. Four samples are randomly picked from each convolution layer.}
    \label{fig:dark_arts}
\end{figure*}
\begin{figure}[h]
    \centering
    \scalebox{0.5}{\includegraphics[width=\textwidth]{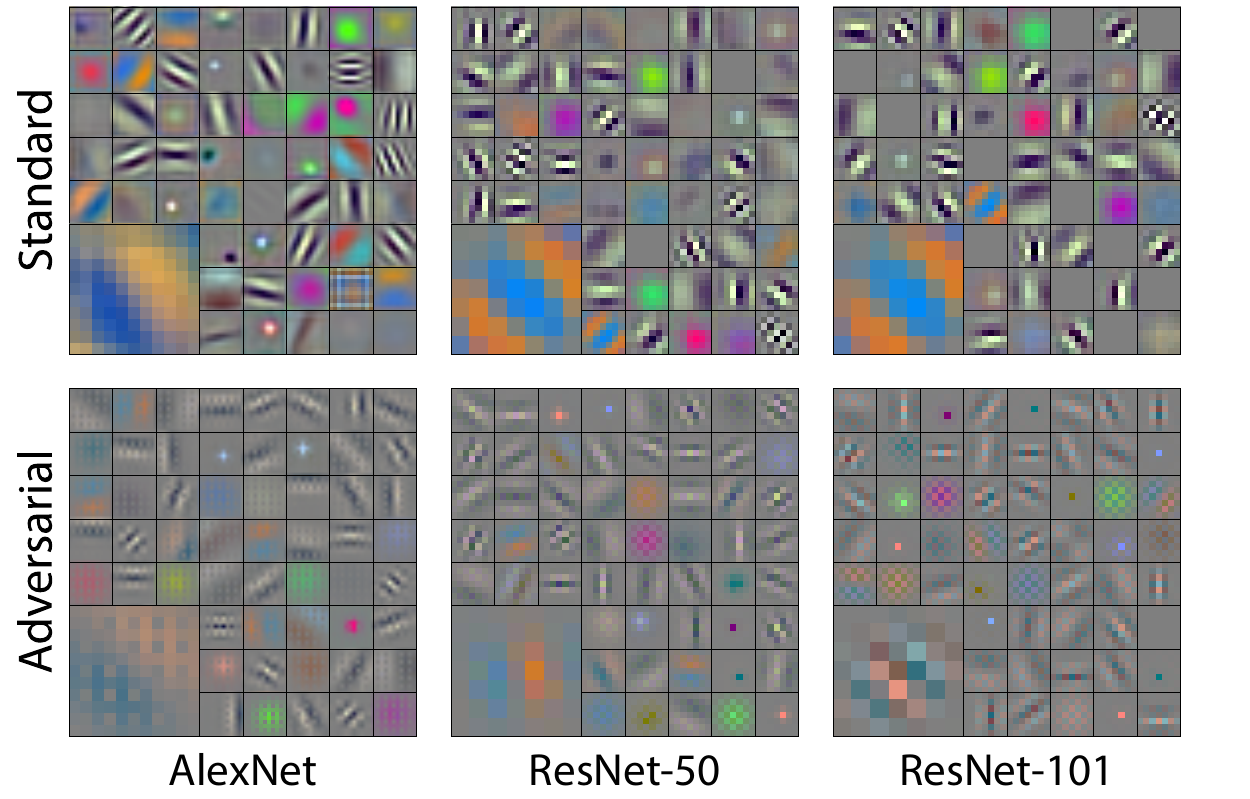}}
    \caption{
        First layer visualization~\cite{erhan2009visualizing} of the standard trained (top row) and the adversarially trained (bottom row) models on ImageNet.
    }
    \label{fig:first_layer}
\end{figure}
{
\renewcommand{\tablename}{Figure}
\setcounter{table}{4}

\begin{table}[t]
    \centering
    \begin{tabular}{ccccc}
        \begin{minipage}{0.13\textwidth}
            \centering
            \scalebox{0.3}{\includegraphics{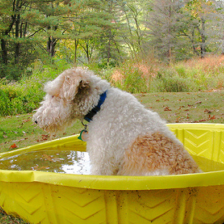}}\\
            Input\\
            \vspace{1.5mm}
        \end{minipage}&
        \rotatebox[origin=c]{90}{\,\,\,\,\,\,\,\,Gaussian Blur}
        \begin{minipage}{0.13\textwidth}
            \centering
            \scalebox{0.3}{\includegraphics{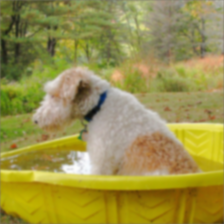}}\\
            $\sigma=1.0$\\
            \vspace{2mm}
        \end{minipage}&
        \begin{minipage}{0.13\textwidth}
            \centering
            \scalebox{0.3}{\includegraphics{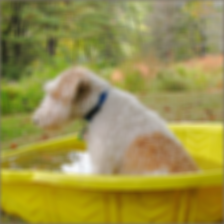}}\\
            $\sigma=3.0$\\
            \vspace{2mm}
        \end{minipage}\\
        
        &
        \rotatebox[origin=c]{90}{\,\,\,\,\,\,Median Filter}
        \begin{minipage}{0.13\textwidth}
            \centering
            \scalebox{0.3}{\includegraphics{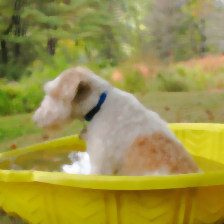}}\\
            $1$ time
        \end{minipage}&
        \begin{minipage}{0.13\textwidth}
            \centering
            \scalebox{0.3}{\includegraphics{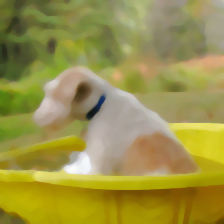}}\\
            $5$ times
        \end{minipage}\\
    \end{tabular}
    \caption{Example images that are applied with Gaussian blur and median filter.}
    \label{fig:blur_median_examples}
\end{table}

\setcounter{table}{1}
\setcounter{figure}{5}
}
\clearpage
\clearpage
\begin{figure}[H]
    \centering
    \begin{subfigure}[b]{0.49\textwidth}
        \begin{subfigure}[b]{0.49\textwidth}
            \centering
            \includegraphics[width=\textwidth]{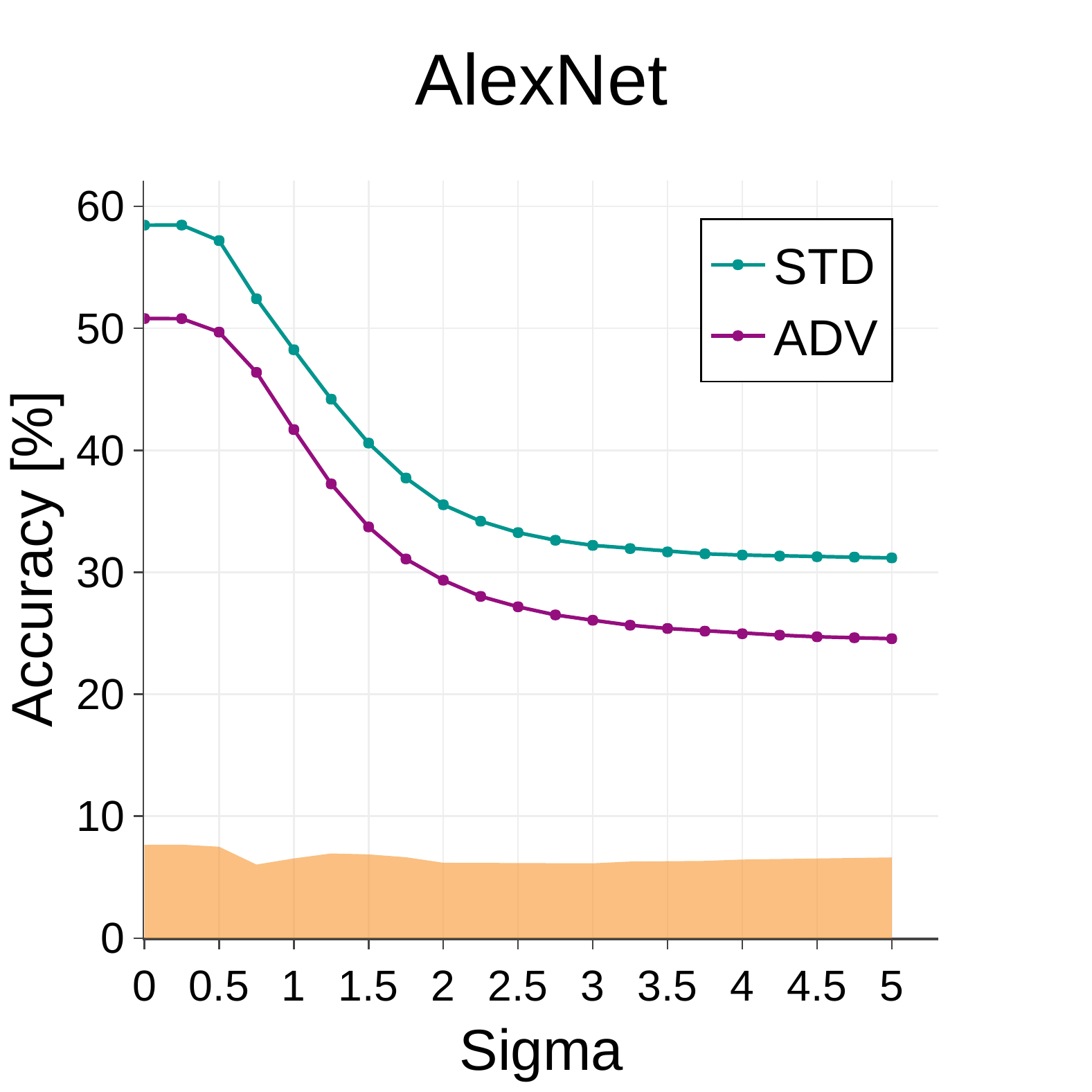}
        \end{subfigure}
        \begin{subfigure}[b]{0.49\textwidth}
            \centering
            \includegraphics[width=\textwidth]{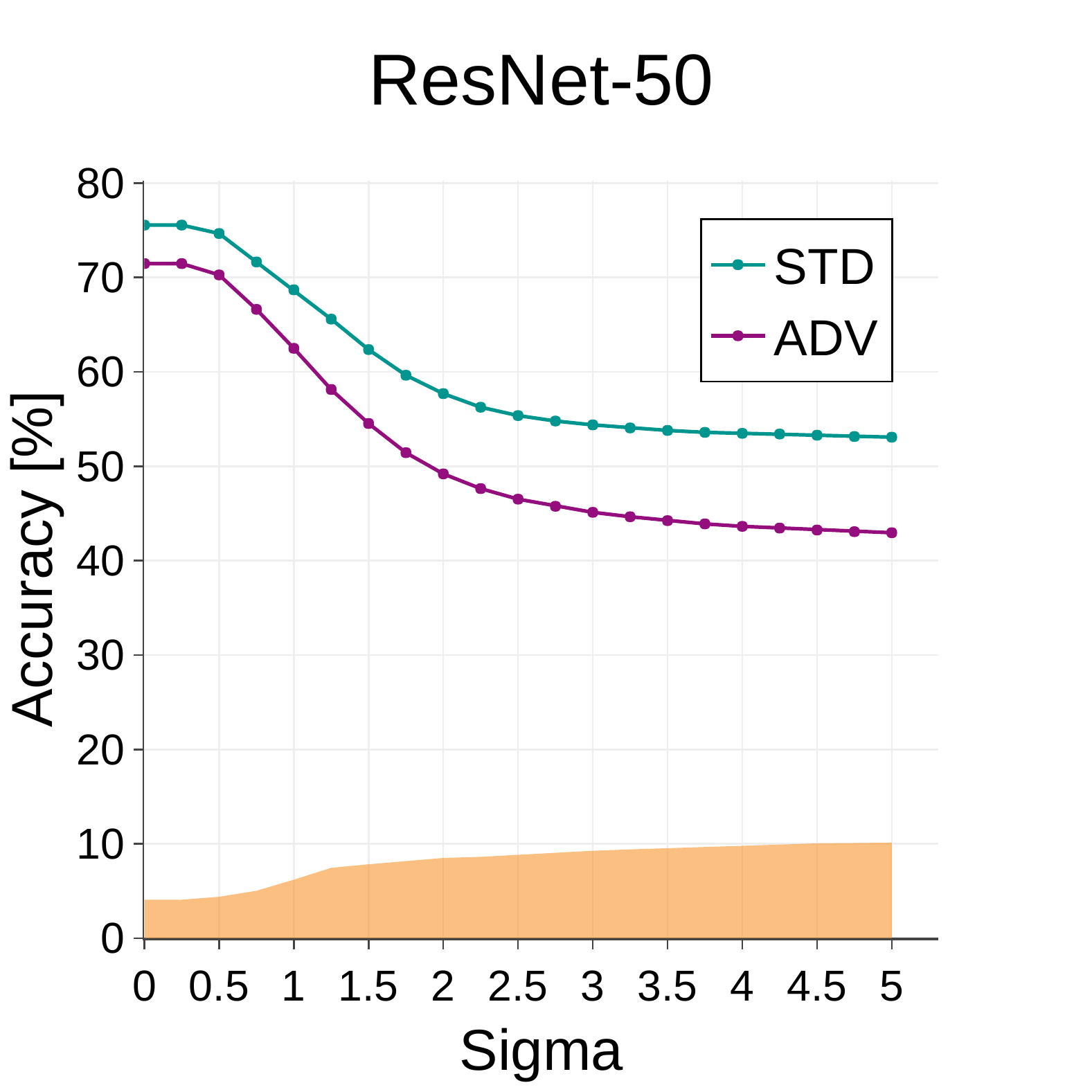}
        \end{subfigure}
        \caption{\normalsize Gaussian Blur}
    \end{subfigure}
    \begin{subfigure}[b]{0.49\textwidth}
        \begin{subfigure}[b]{0.49\textwidth}
            \centering
            \includegraphics[width=\textwidth]{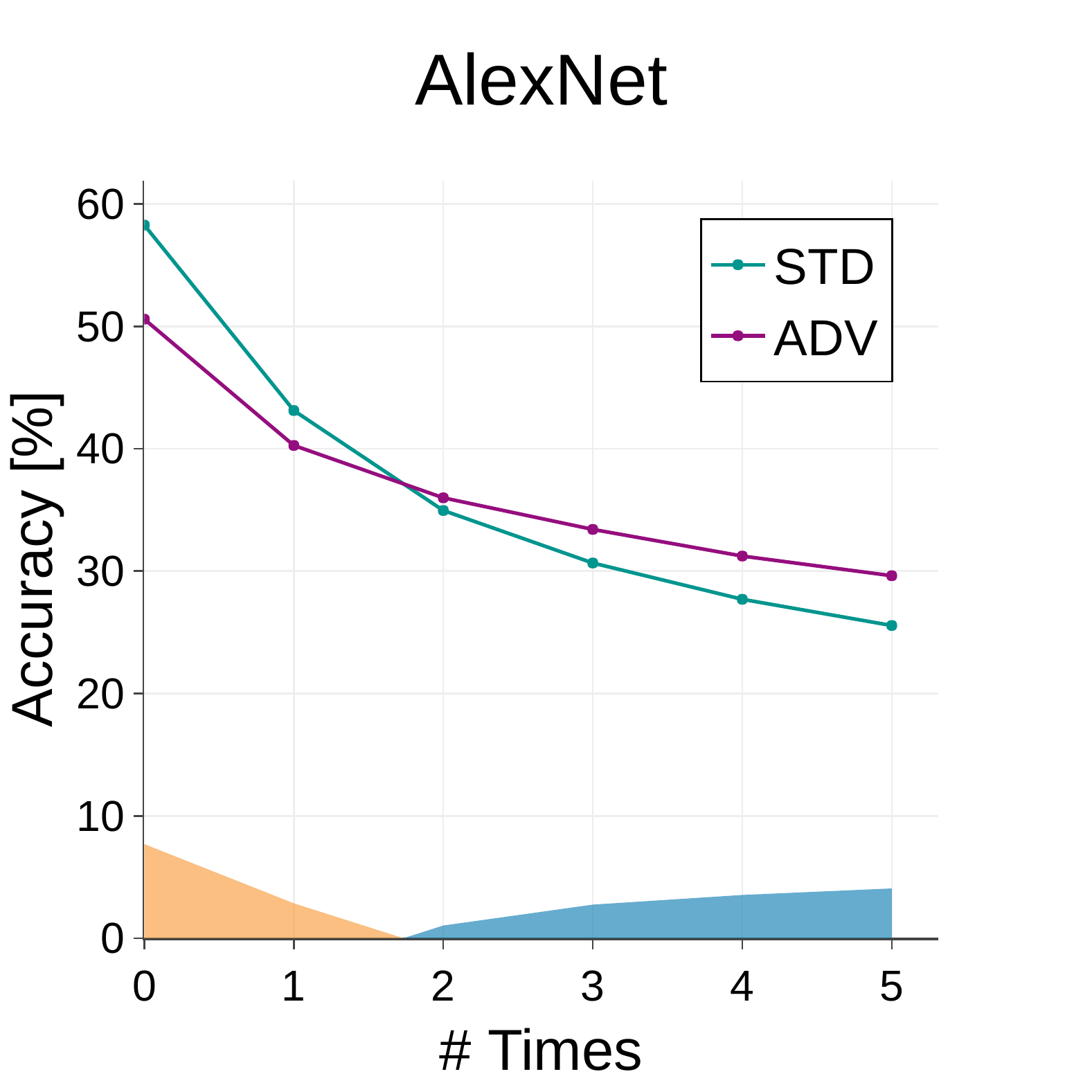}
        \end{subfigure}
        \begin{subfigure}[b]{0.49\textwidth}
            \centering
            \includegraphics[width=\textwidth]{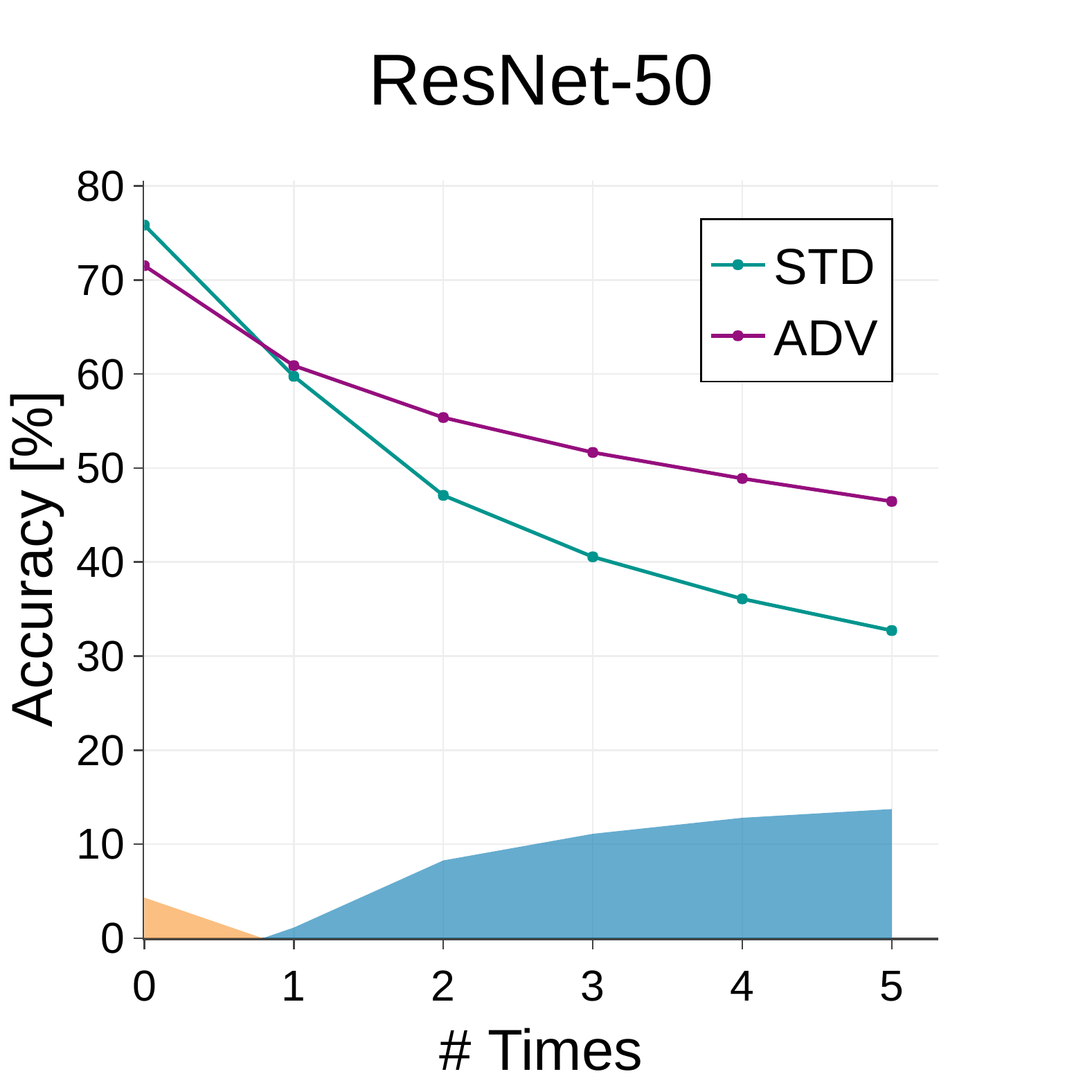}
        \end{subfigure}
        \caption{\normalsize Median Filter}
    \end{subfigure}
    \caption{Results of standard accuracy on ImageNet applied with Gaussian blur and median filter.
    Filled regions mean the difference between standard trained models and adversarially trained models ({\color[rgb]{0.9725,0.5843,0.1804}{Orange}} means that standard trained models exceed and {\color[rgb]{0.0000,0.4627,0.6863}{blue}} means that adversarially trained models exceed).
    Both models are trained on standard ImageNet ({\bf without fine-tuning}).}
    \label{fig:blur_median}
\end{figure}

\subsection{What Do Adversarially Robust Features See?}
\label{subsec:what_do}
To visualize the differences between standard accurate and adversarially robust features, we calculated the sensitivity maps of both the standard trained and the adversarially trained models by Vanilla Grad, Loss Grad and Guided Backprop and the results of ResNet-50 are shown in Figure~\ref{fig:map}.
Vanilla and Loss Grad of the adversarial trained model are more perceivable for humans as previous works reported.
On the other hand, in the results of Guided Backprop, both models are correctly capturing the features of the main objects in the input images, but it can be seen that the adversarially trained model does not respond to the fine texture (e.g., the wing of the insect and the hair of the dog).

Next, we performed layer visualization and the results are shown in Figure~\ref{fig:dark_arts} and 
Figure~\ref{fig:first_layer}.
It can be confirmed that the values of the first layers of the adversarially trained models are smaller than that of the standard trained ones to be insensitive to weak edges and textures.
The deeper layers of the standard trained model have the small color cluttering that cannot be confirmed in those of the adversarially trained model.
Furthermore, it can be seen that the adversarially trained model captures an obviously larger structure than the standard trained model.

In order to confirm this fact quantitatively, we added some artificial transformations to the validation set of ImageNet and investigated changes in accuracy of the standard trained and the adversarially trained models.
As artificial transformations, we applied Gaussian blur and median filter to reduce the spatial information, color reversal to modify the color information and stylization by style transfer techniques to change the texture while keeping the edge and silhouette information as Geirhos et al.~\cite{geirhos2018imagenet} did.
In the Gaussian blur setting, we set the kernel size to $7$ and changed the value of $\sigma$ from $0$ to $5$ in increments of $0.25$.
In the median filter setting, we set the kernel size to $5$ and changed the number of times to apply the median filter from $0$ to $5$.

The example images and the results of the Gaussian blur and the median filter settings are represented in Figure~\ref{fig:blur_median_examples} and Figure~\ref{fig:blur_median}, respectively.
In the Gaussian blur setting, the accuracy of both models decreases while maintaining the difference of their accuracy.
On the other hand, in the median filter setting, the adversarially trained model outperform the standard trained one as the times of median filter increases.
This tendency can be confirmed in both AlexNet and ResNet-50.
This results suggest that the adversarially trained model uses stronger edges such as the silhouette information of objects rather than the detailed texture information, since the Gaussian blur makes all edges blurry equally while the median filter only removes weak edges and noise.

\begin{table*}[t]
    \centering
    \begin{tabular}{c|c|c|p{6em}|p{6em}|p{6em}} \Hline
        \multirow{2}{*}{Architecture} & \multirow{2}{*}{Setting} & \multirow{2}{*}{Metric} & \multicolumn{3}{c}{Variants of ImageNet} \\ \cline{4-6}
        & & & \hfil Standard \hfil & \hfil Reversed \hfil & \hfil Stylized \hfil \\ \Hline
        \multirow{4}{*}{AlexNet} 
            & \multirow{2}{*}{STD} 
                & Top-1 Acc [\%] & \hfil $\mathbf{55.28} \pm \mathbf{0.18}$ \hfil & \hfil $\mathbf{13.18} \pm \mathbf{0.17}$ \hfil & \hfil $4.20 \pm 0.06$ \hfil \\
            &   & Top-5 Acc [\%] & \hfil $\mathbf{80.60} \pm \mathbf{0.05}$ \hfil & \hfil $\mathbf{28.88} \pm \mathbf{0.35}$ \hfil & \hfil $10.74 \pm 0.15$ \hfil \\ \cline{2-6}
            & \multirow{2}{*}{ADV}
                & Top-1 Acc [\%] & \hfil $50.59 \pm 0.22$ \hfil & \hfil $7.97 \pm 0.11$ \hfil & \hfil $\mathbf{5.07} \pm \mathbf{0.04}$ \hfil \\
            &   & Top-5 Acc [\%] & \hfil $73.73 \pm 0.07$ \hfil & \hfil $19.09 \pm 0.23$ \hfil & \hfil $\mathbf{11.86} \pm \mathbf{0.10}$ \hfil \\ \Hline
        \multirow{4}{*}{ResNet-50}
            & \multirow{2}{*}{STD} 
                & Top-1 Acc [\%] & \hfil $\mathbf{75.84} \pm \mathbf{0.31}$ \hfil & \hfil $\mathbf{32.36} \pm \mathbf{0.77}$ \hfil & \hfil $6.90 \pm 0.22$ \hfil \\
            &   & Top-5 Acc [\%] & \hfil $\mathbf{92.74} \pm \mathbf{0.14}$ \hfil & \hfil $\mathbf{56.61} \pm \mathbf{0.87}$ \hfil & \hfil $15.52 \pm 0.23$ \hfil \\ \cline{2-6}
            & \multirow{2}{*}{ADV}
                & Top-1 Acc [\%] & \hfil $71.52 \pm 0.06$ \hfil & \hfil $23.26 \pm 0.14$ \hfil & \hfil $\mathbf{11.08} \pm \mathbf{0.26}$ \hfil \\
            &   & Top-5 Acc [\%] & \hfil $90.02 \pm 0.03$ \hfil & \hfil $43.01 \pm 0.42$ \hfil & \hfil $\mathbf{21.89} \pm \mathbf{0.32}$ \hfil \\ \Hline
    \end{tabular}
    \caption{
        Results of standard accuracy on variants of ImageNet. 
        All the models are trained on standard ImageNet ({\bf without fine-tuning}).
        Reversed variant reverses pixel values around mean values of ImageNet ($[R,G,B]=[0.485, 0.456, 0.406]$). 
        Stylized variant uses Stylized ImageNet~\cite{geirhos2018imagenet}. 
        Example images of each variant are shown in the supplementary materials.
    }
    \label{tab:stylize}
\end{table*}
\begin{table}[!ht]
    \centering
    \begin{tabular}{c|c|c|c} \Hline
    Dataset & \# Pixel & \# Class & \# Train \\ \Hline
    SVHN~\cite{svhn} & $32$ & $10$ & $73,257$ \\
    CIFAR-10~\cite{cifar} & $32$ & $10$ & $50,000$ \\
    CIFAR-100~\cite{cifar} & $32$ & $100$ & $50,000$ \\
    Tiny ImageNet~\cite{tiny_imagenet} & $64$ & $200$ & $100,000$ \\
    STL-10~\cite{stl10} & $96$ & $10$ & $5,000$ \\
    Food-101~\cite{food101} & $\approx 549$ & $101$ & $75,750$ \\
    Flower-102~\cite{flower102} & $\approx 583$ & $102$ & $1,020$ \\
    Stanford Dogs~\cite{stanforddogs} & $\approx 414$ & $120$ & $12,000$ \\
    CUB-200~\cite{cub200} & $\approx 426$ & $200$ & $5,994$ \\
    ImageNet~\cite{imagenet} & $\approx 460$ & $1,000$ & $1,281,167$ \\
    \Hline
    \end{tabular}
    \caption{Information of datasets for image classification. The number of pixels are the average values of all the images of the validation sets.}
    \label{tab:datasets}
\end{table}

Next, Table~\ref{tab:stylize} shows the result of applying color reversal and style transfer.
Although, in color reversal setting, the same range of accuracy reductions were observed in both models, the accuracy of adversarial trained model exceeded standard trained one in Stylized ImageNet.
Since Stylized ImageNet stylizes the texture inside the objects, it suggests that the adversarially trained models make shape based decisions rather than texture comparing to the standard trained models.
Following the fact that humans are able to achieve high accuracy using only the silhouette information as reported by Geirhos et al.~\cite{geirhos2018imagenet}, these results follows the report of Tsipras et al.~\cite{tsipras2019robustness} that adversarially trained models are better align to human perception than standard trained ones.

\subsection{Benefits of Adversarially Robust Features}
As described above, standard accurate and adversarially robust models obtain different sets of features.
This result induces our next simple question: ``Can adversarially robust features be utilized to improve accuracy by combining with standard accurate features?''
To answer this question, we performed ensemble learning with combination of standard trained and adversarially trained models as pre-trained models three times for 10 datasets (Table~\ref{tab:datasets}).
We employed the late fusion scheme that merges two classifiers in the last layer.
Then, we trained models for 30 epochs with the learning rate 0.001 and set the other parameters to the same as described above.
For fair comparison, we also trained not only plain models without ensemble learning (STD, ADV) but also the ensemble models of two different standard trained models (STD+STD) and that of two different adversarially trained models (ADV+ADV).

\begin{figure*}[t]
    \centering
    \begin{subfigure}[t]{0.99\textwidth}
        \centering
        \begin{subfigure}[b]{0.19\textwidth}
            \centering
            \includegraphics[width=\textwidth]{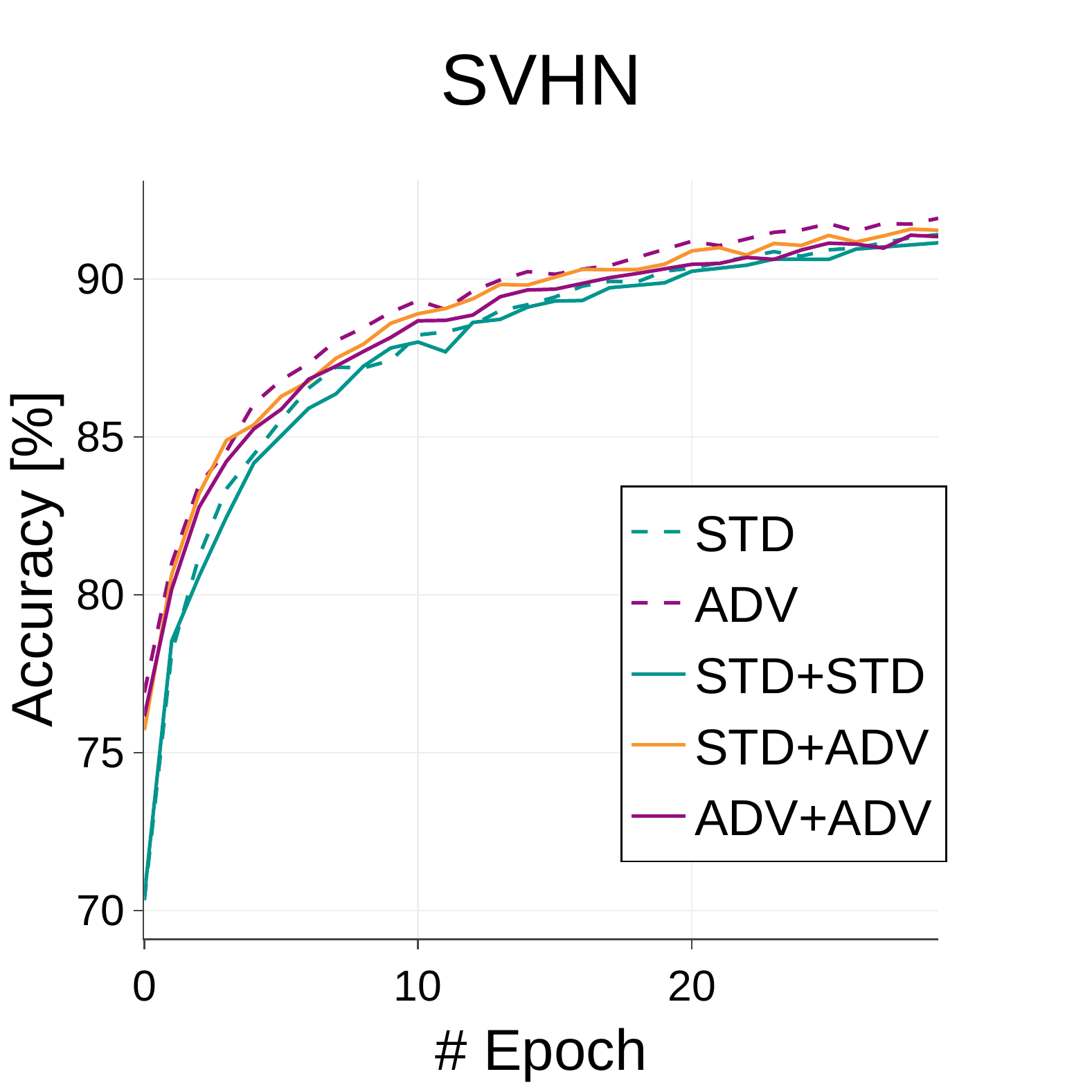}
        \end{subfigure}
        \begin{subfigure}[b]{0.19\textwidth}
            \centering
            \includegraphics[width=\textwidth]{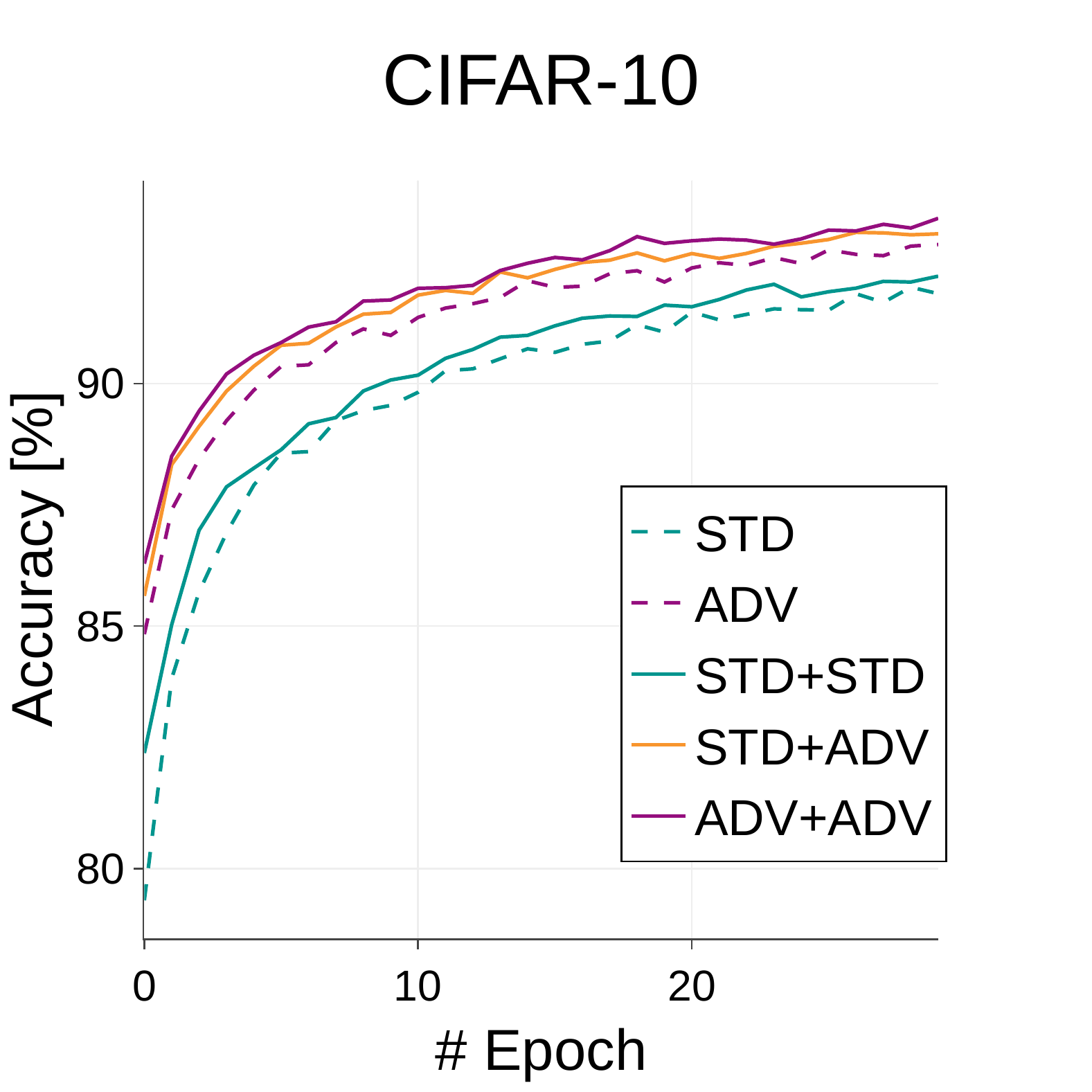}
        \end{subfigure}
        \begin{subfigure}[b]{0.19\textwidth}
            \centering
            \includegraphics[width=\textwidth]{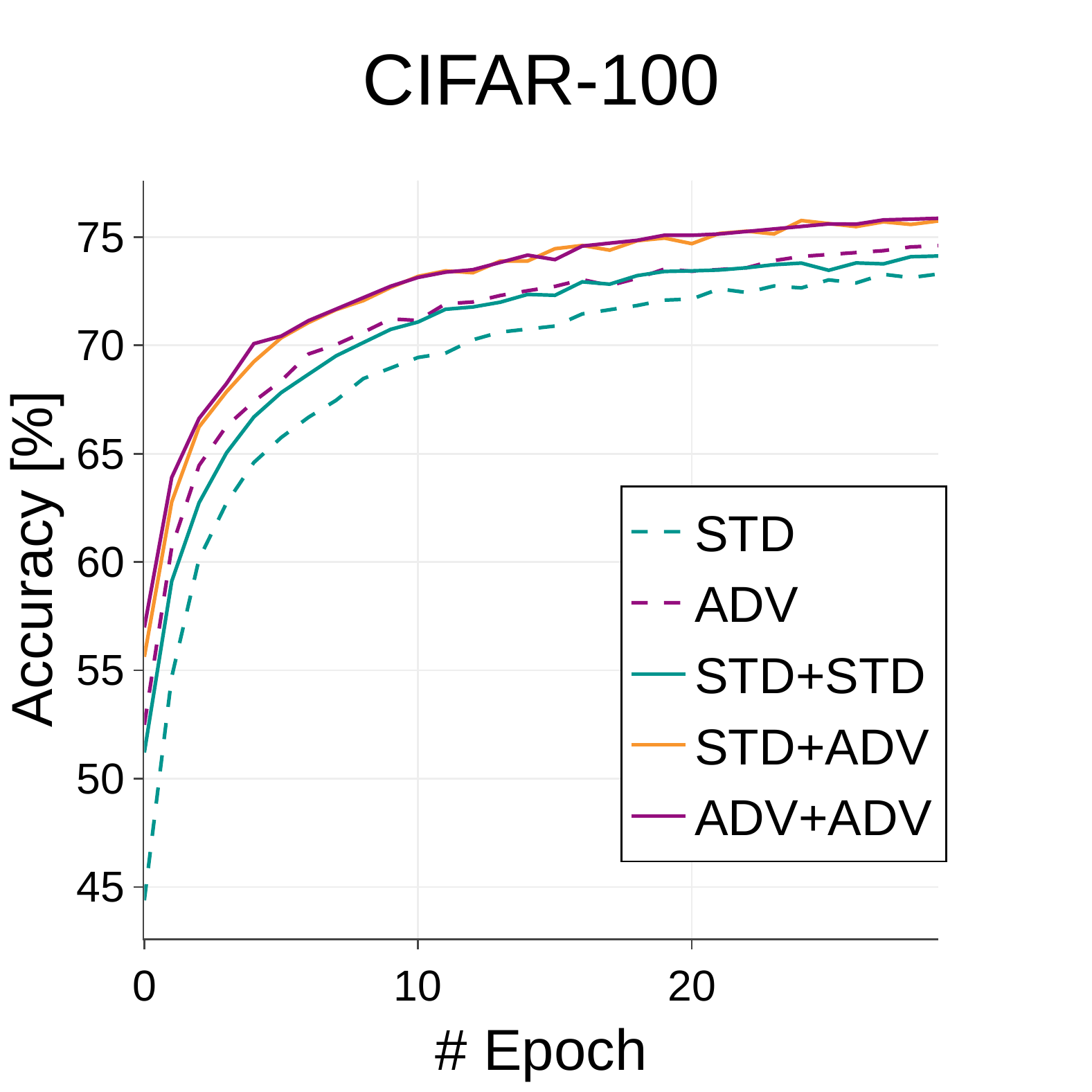}
        \end{subfigure}
        \begin{subfigure}[b]{0.19\textwidth}
            \centering
            \includegraphics[width=\textwidth]{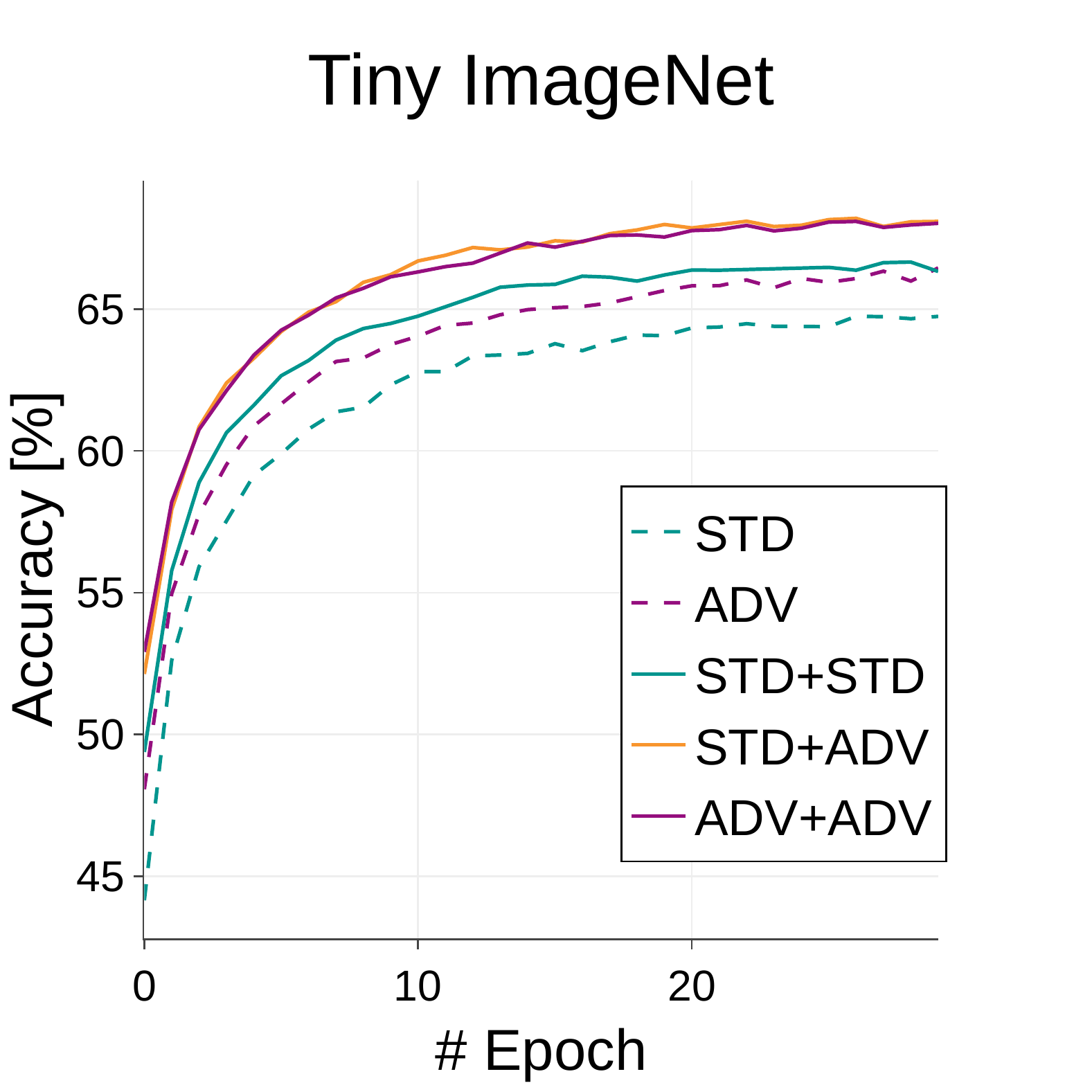}
        \end{subfigure}
        \begin{subfigure}[b]{0.19\textwidth}
            \centering
            \includegraphics[width=\textwidth]{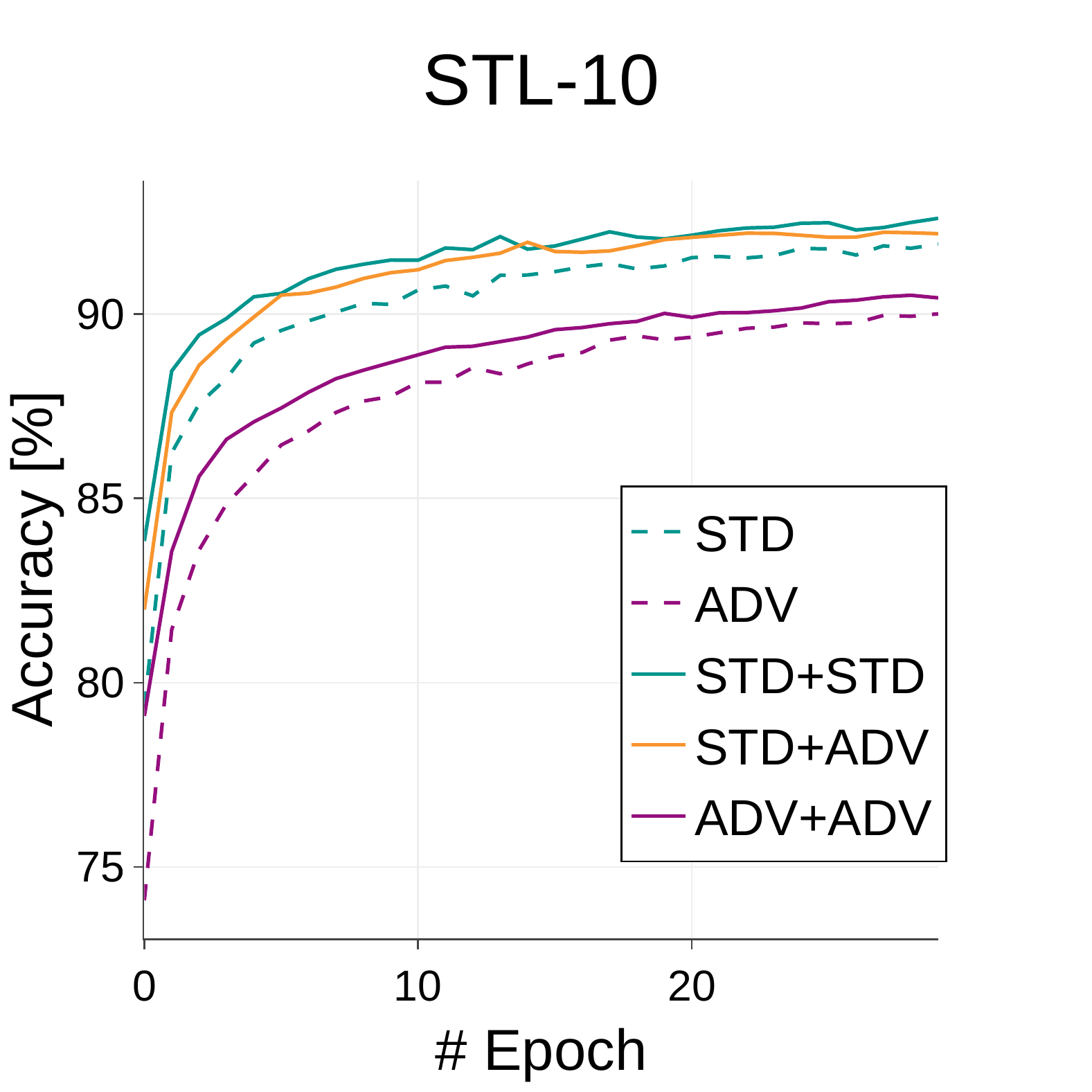}
        \end{subfigure}
        \begin{subfigure}[b]{0.19\textwidth}
            \centering
            \includegraphics[width=\textwidth]{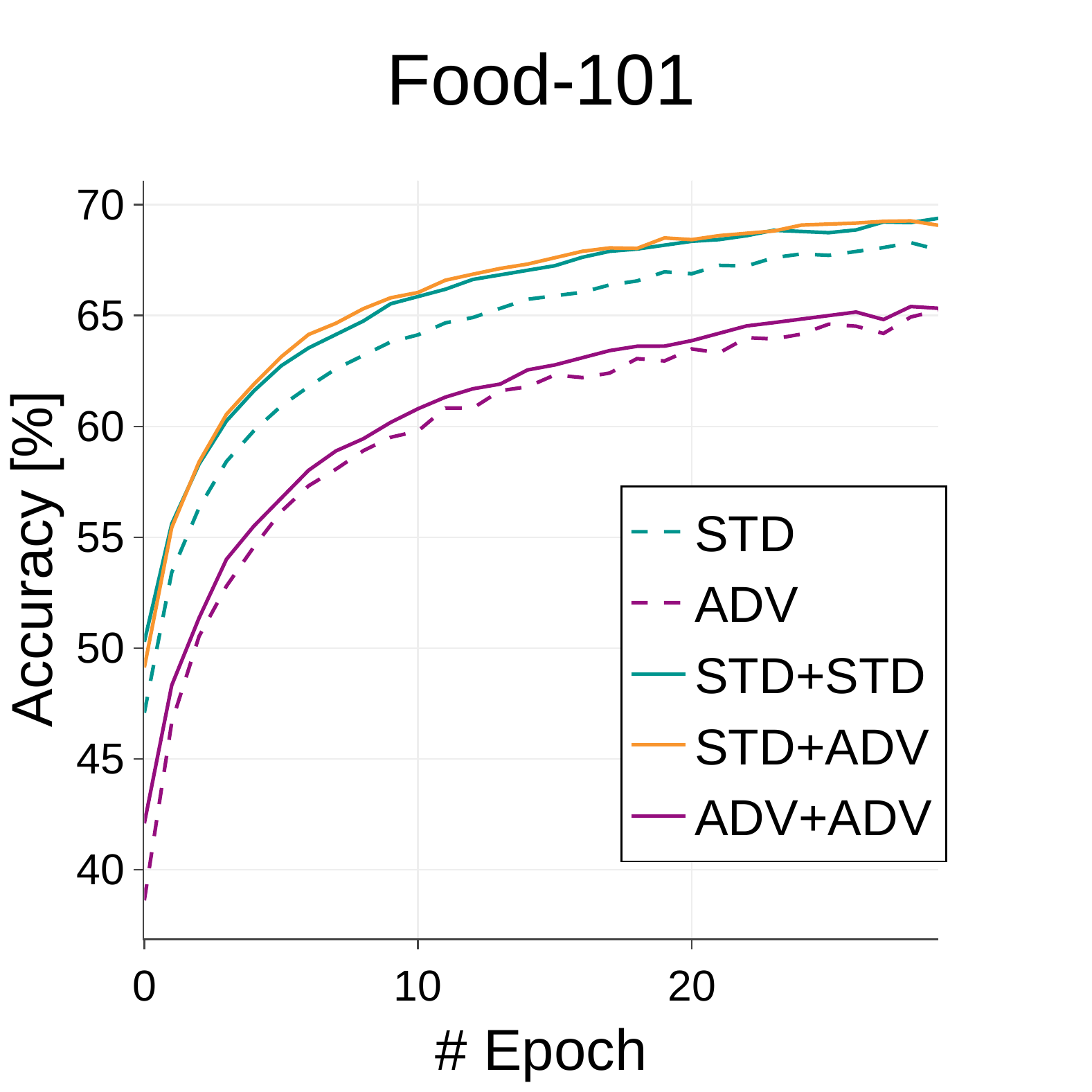}
        \end{subfigure}
        \begin{subfigure}[b]{0.19\textwidth}
            \centering
            \includegraphics[width=\textwidth]{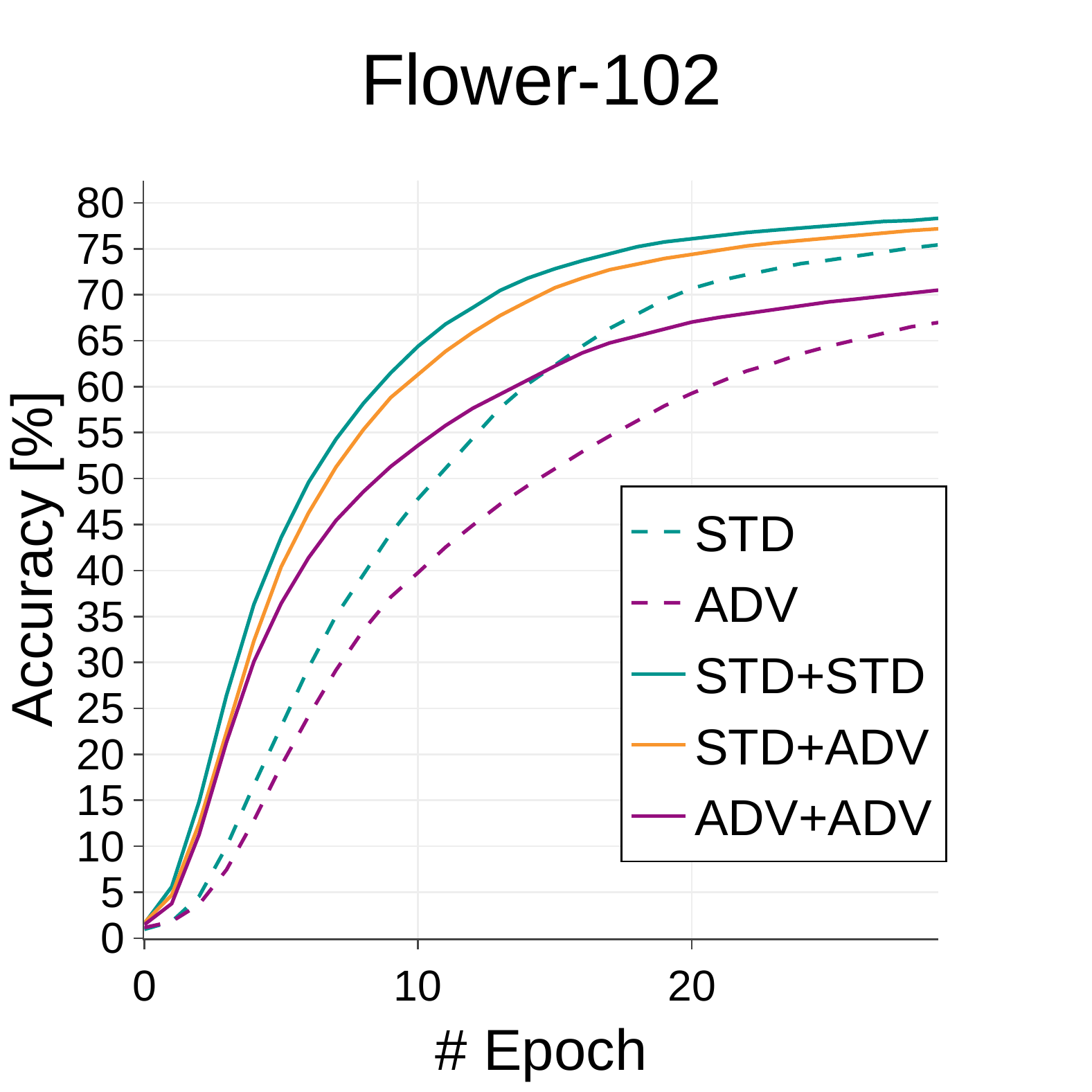}
        \end{subfigure}
        \begin{subfigure}[b]{0.19\textwidth}
            \centering
            \includegraphics[width=\textwidth]{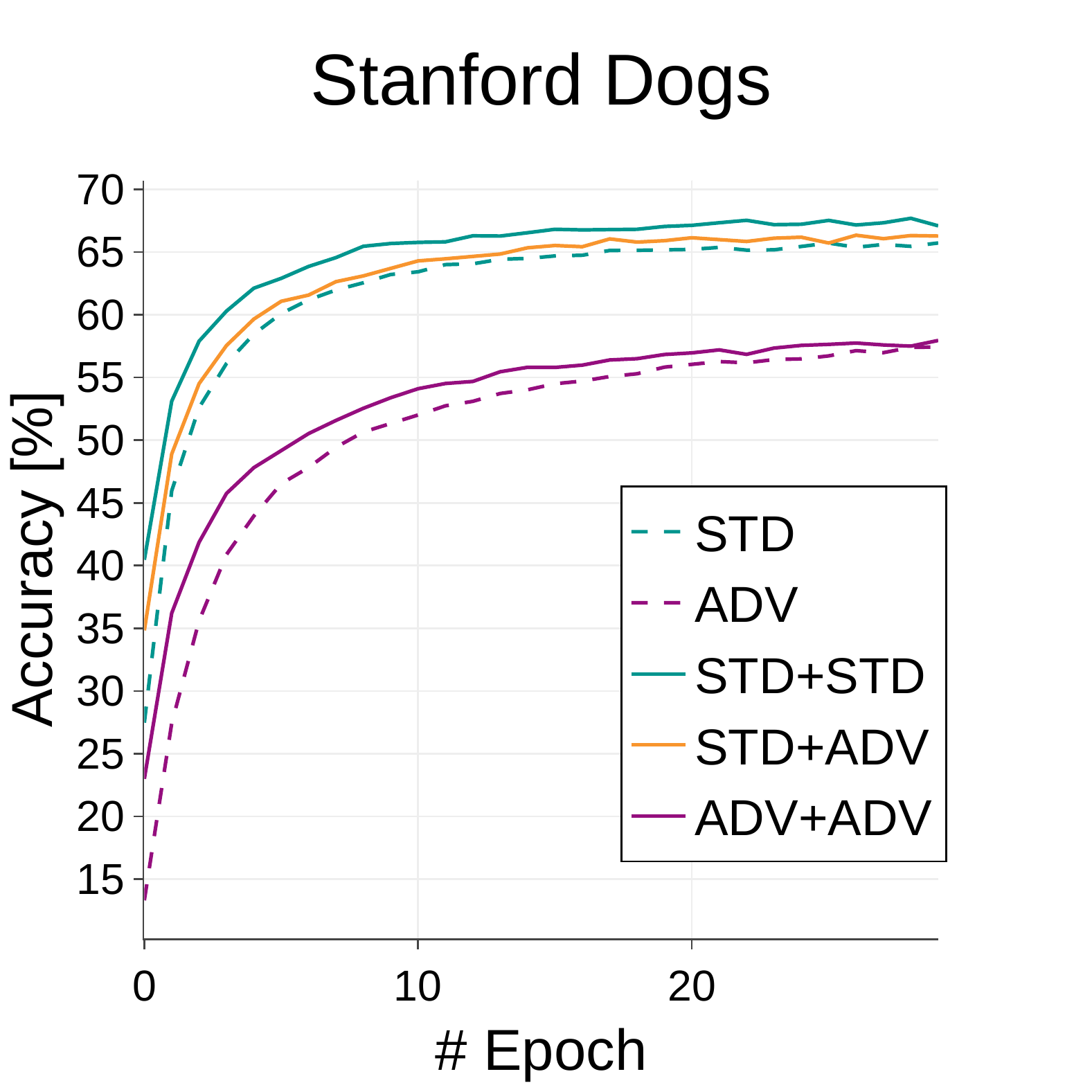}
        \end{subfigure}
        \begin{subfigure}[b]{0.19\textwidth}
            \centering
            \includegraphics[width=\textwidth]{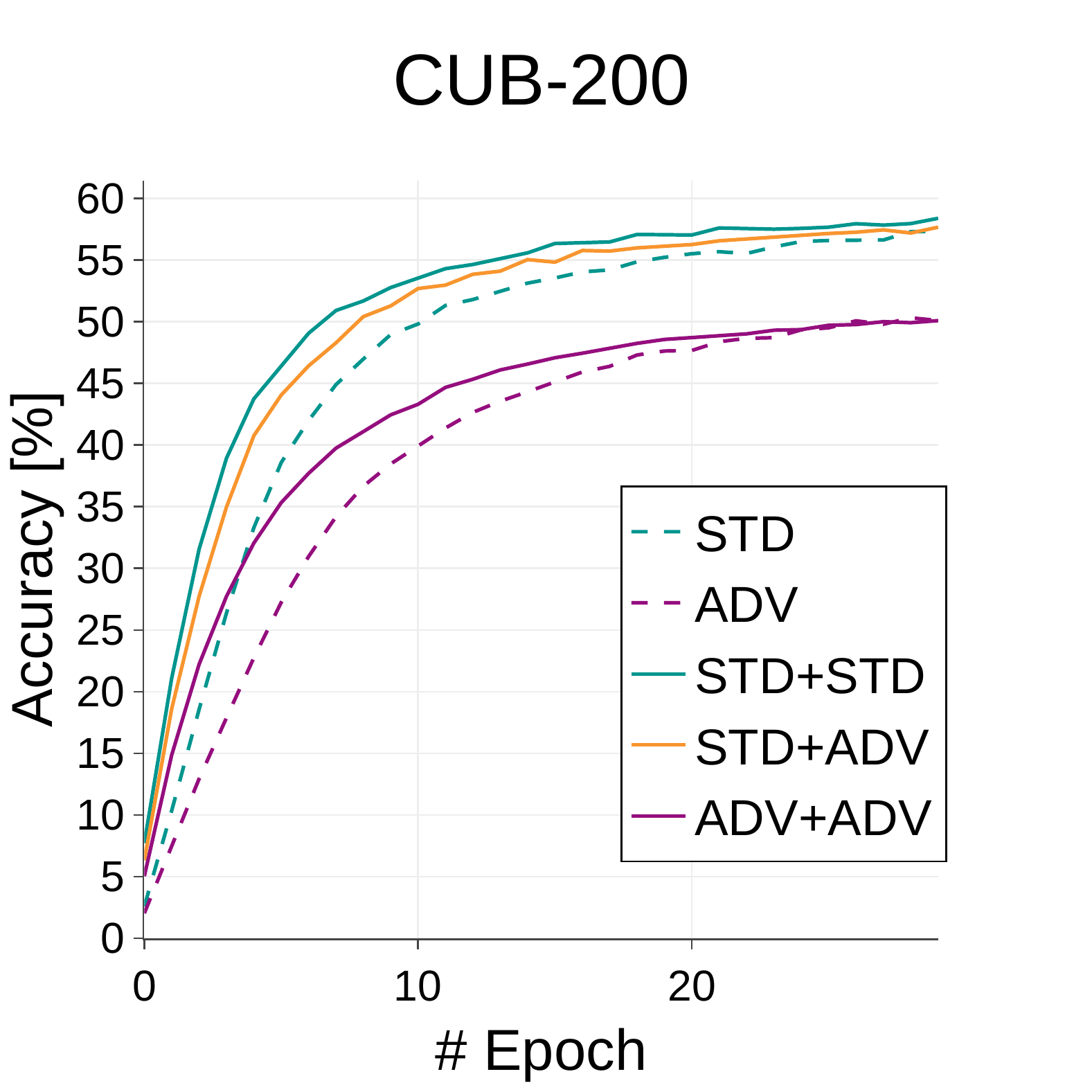}
        \end{subfigure}
        \begin{subfigure}[b]{0.19\textwidth}
            \centering
            \includegraphics[width=\textwidth]{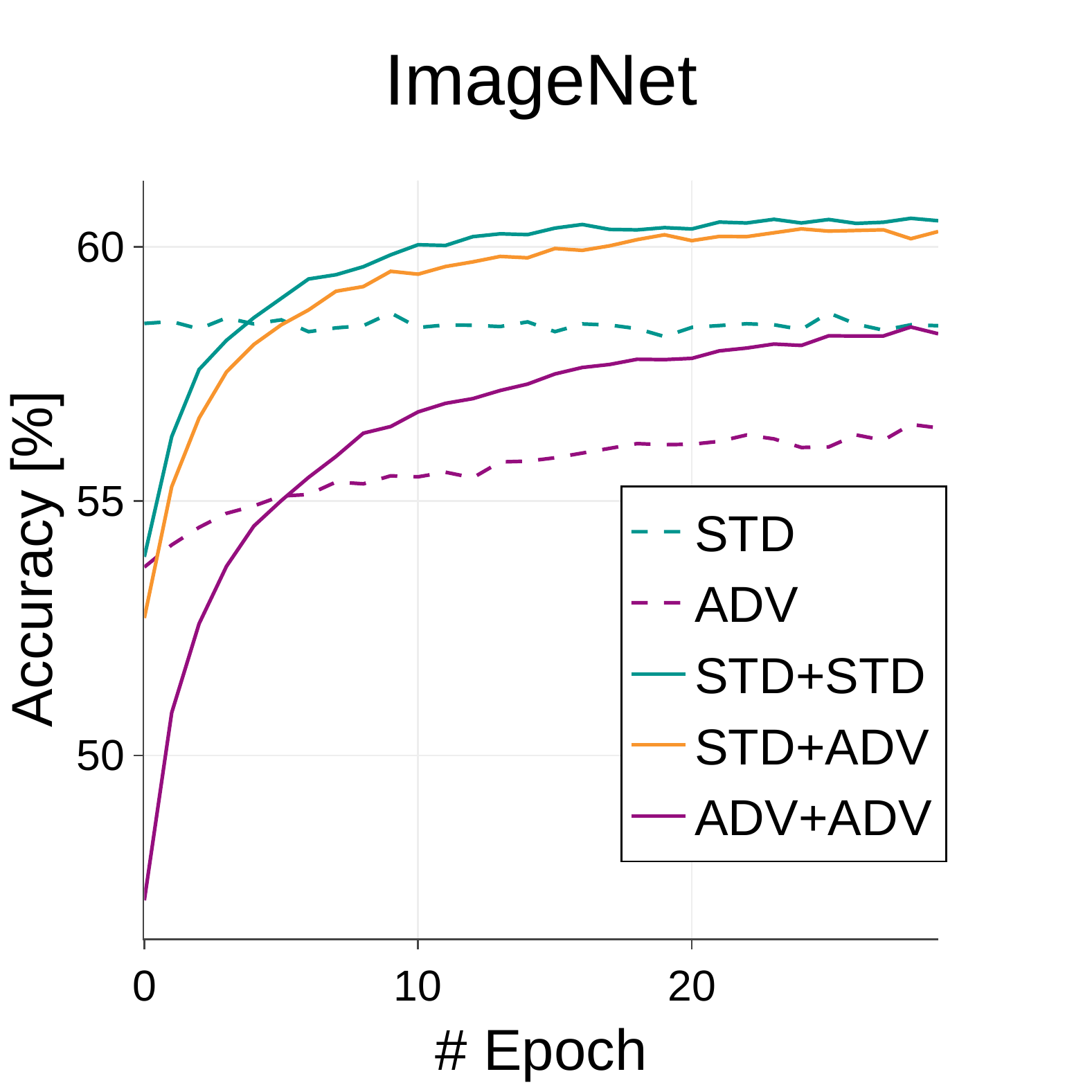}
        \end{subfigure}
        \caption{AlexNet}
    \end{subfigure}
    
    \begin{subfigure}[t]{0.99\textwidth}
        \centering
        \begin{subfigure}[b]{0.19\textwidth}
            \centering
            \includegraphics[width=\textwidth]{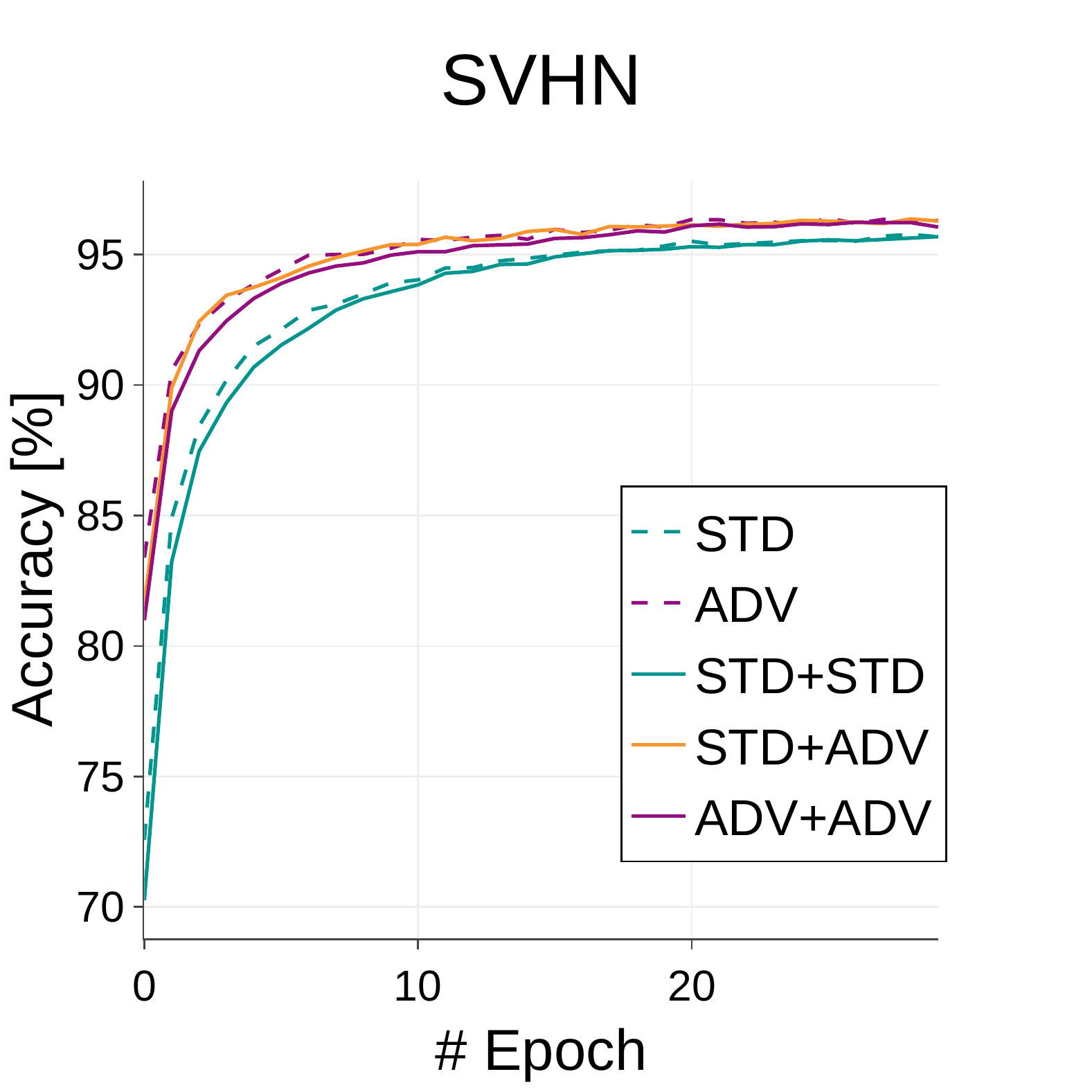}
        \end{subfigure}
        \begin{subfigure}[b]{0.19\textwidth}
            \centering
            \includegraphics[width=\textwidth]{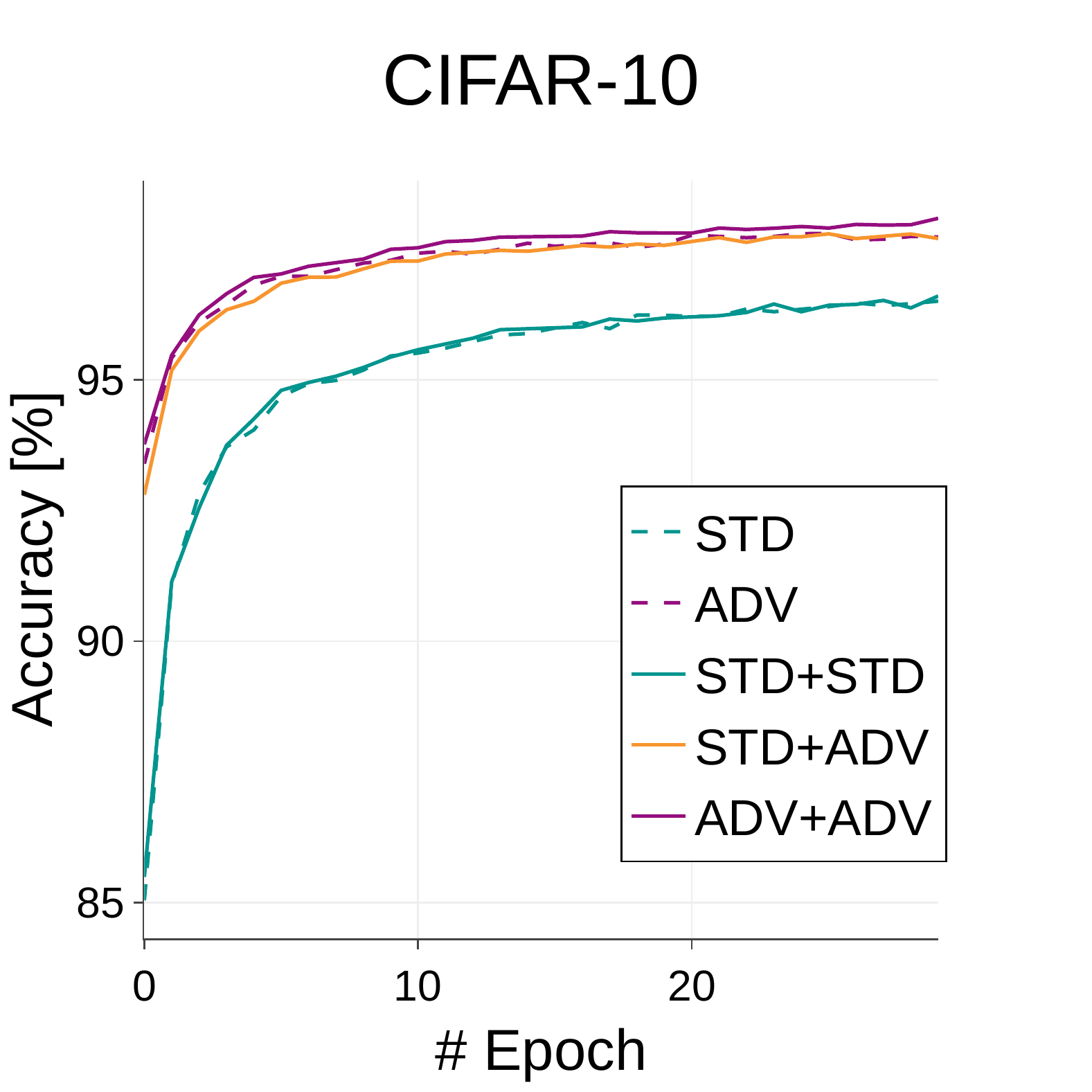}
        \end{subfigure}
        \begin{subfigure}[b]{0.19\textwidth}
            \centering
            \includegraphics[width=\textwidth]{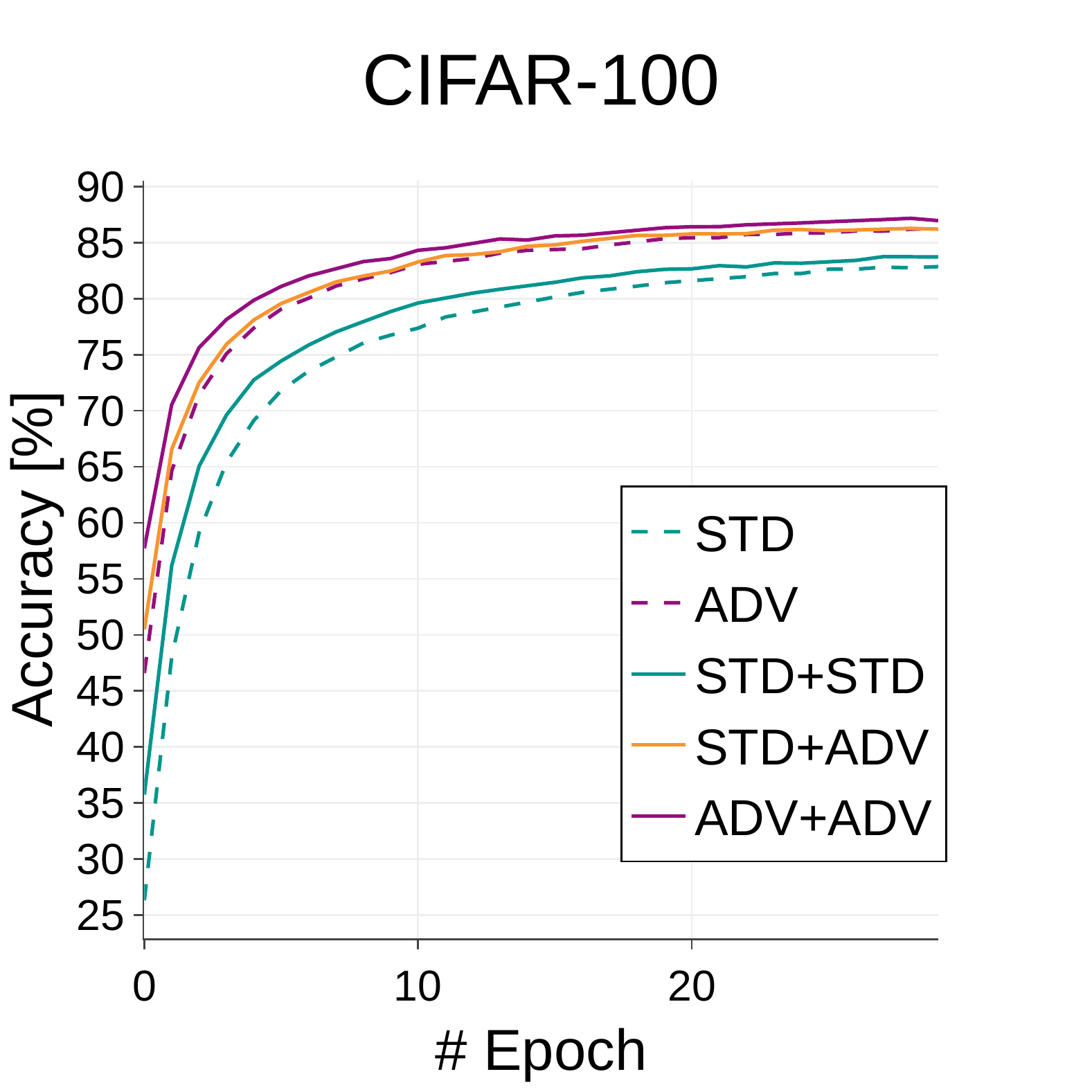}
        \end{subfigure}
        \begin{subfigure}[b]{0.19\textwidth}
            \centering
            \includegraphics[width=\textwidth]{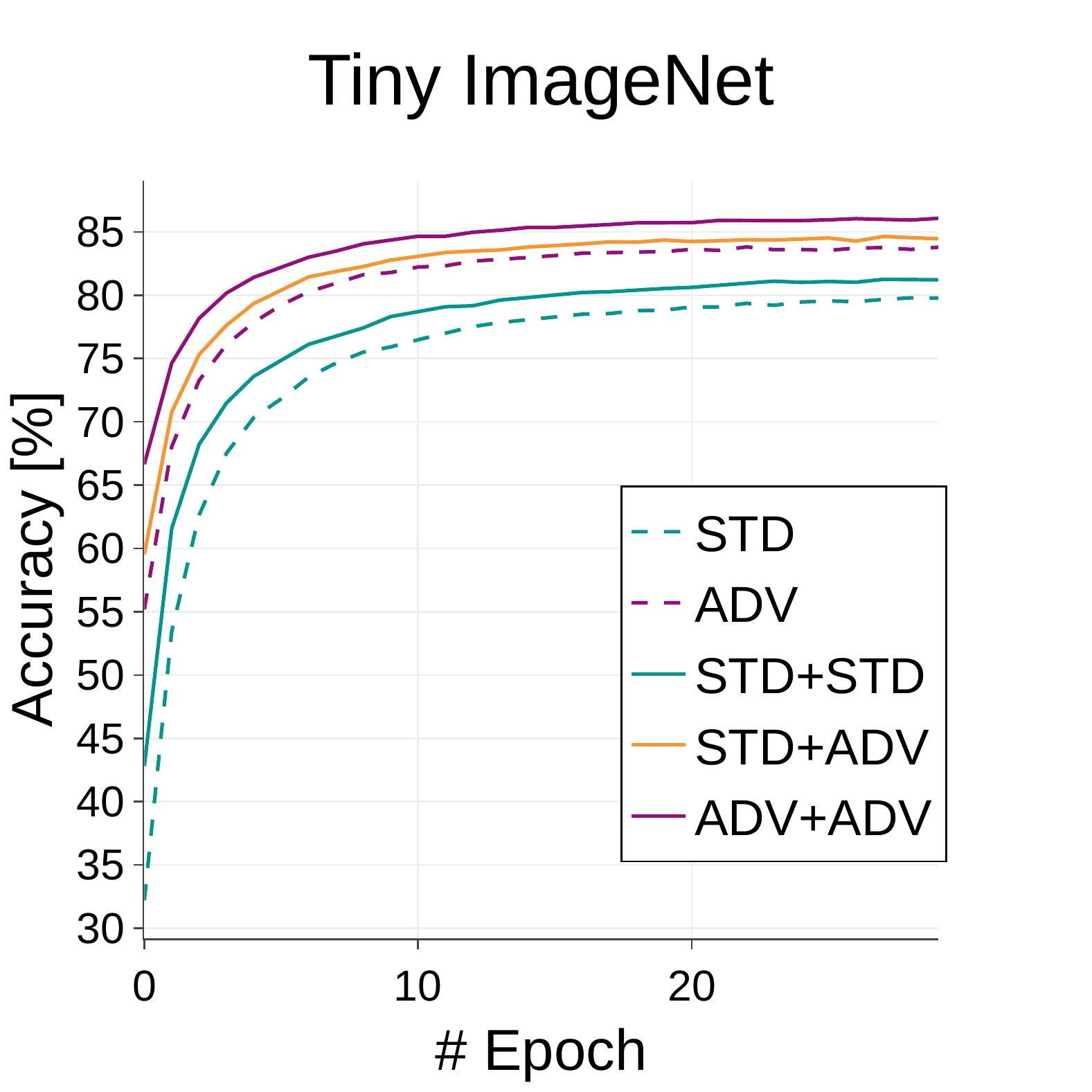}
        \end{subfigure}
        \begin{subfigure}[b]{0.19\textwidth}
            \centering
            \includegraphics[width=\textwidth]{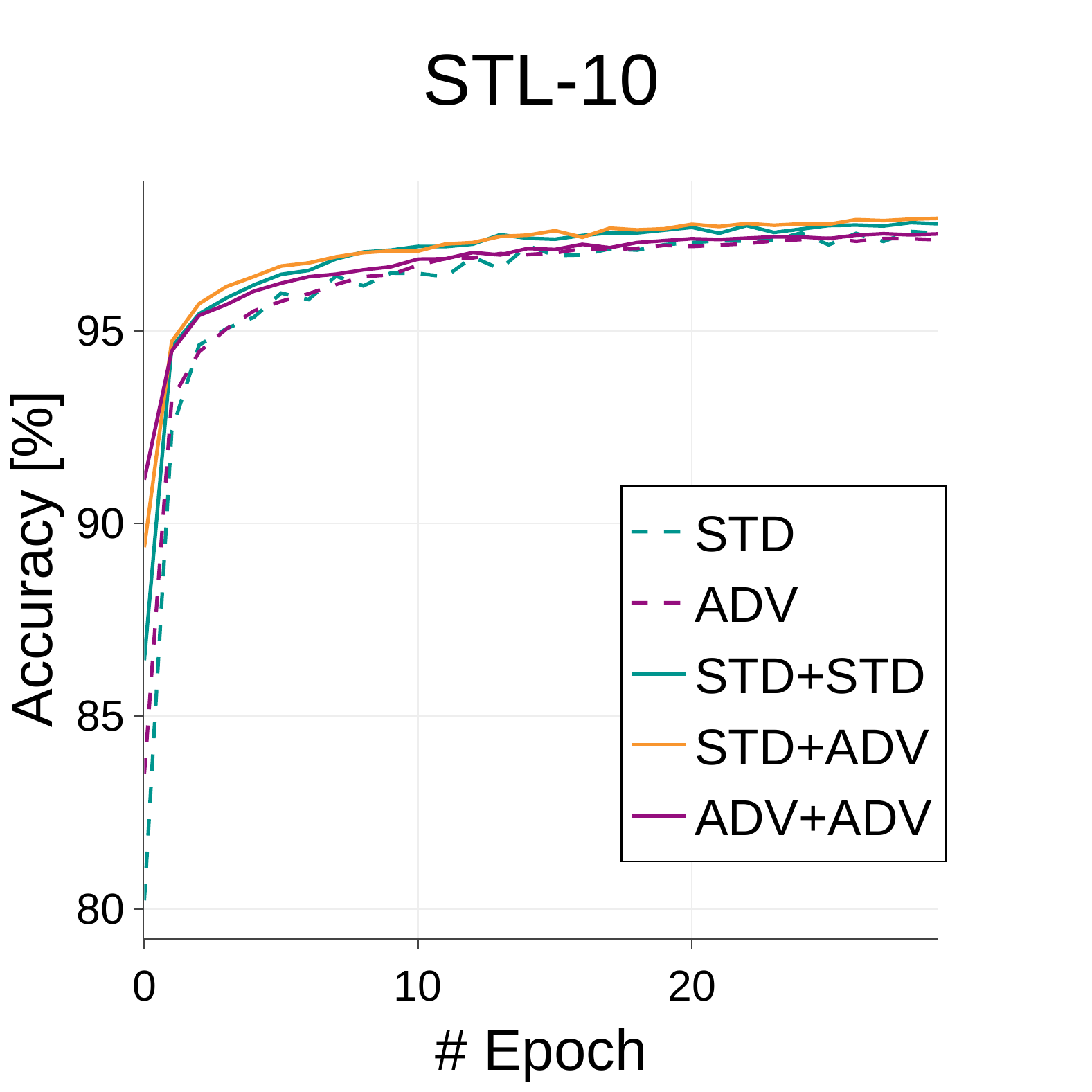}
        \end{subfigure}
        \begin{subfigure}[b]{0.19\textwidth}
            \centering
            \includegraphics[width=\textwidth]{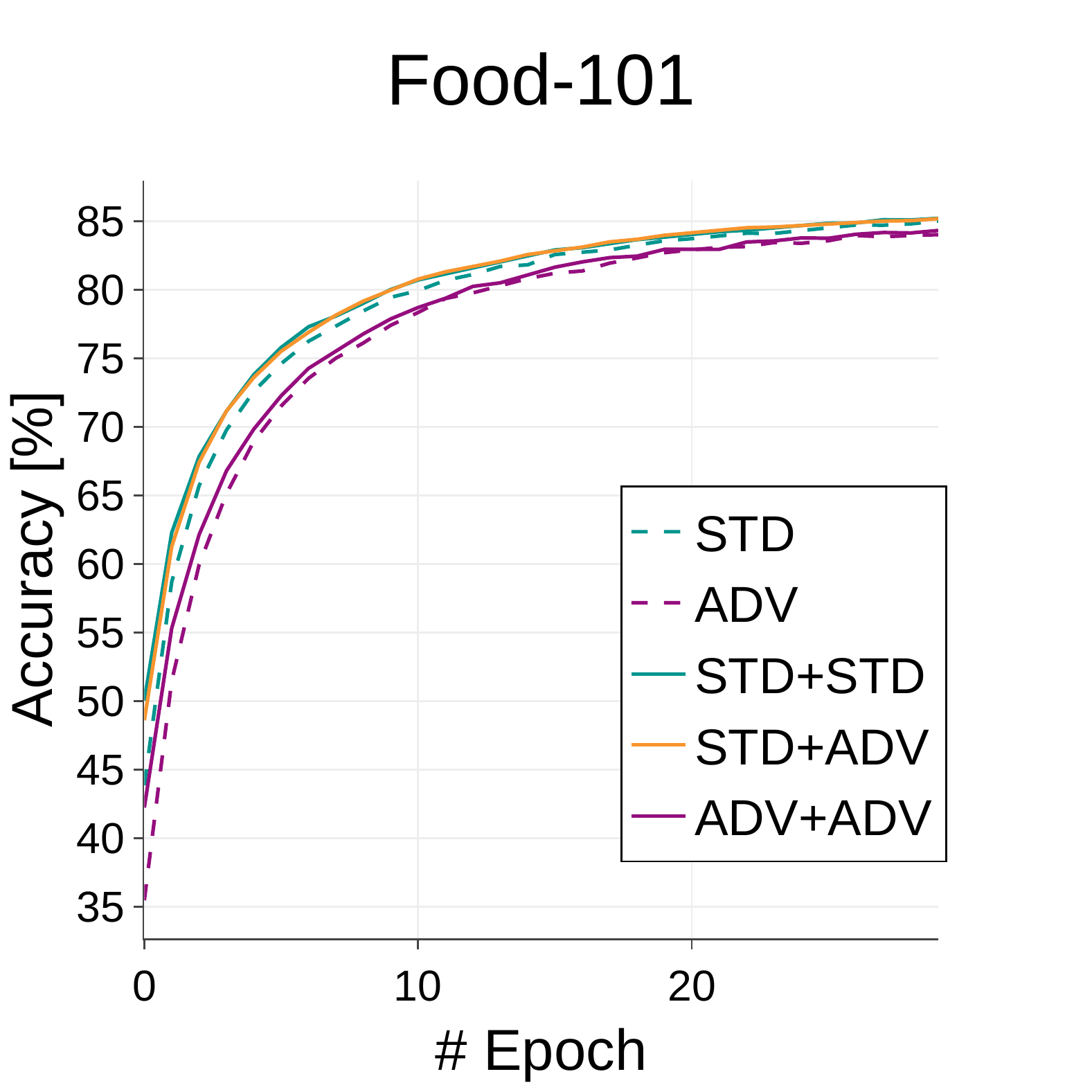}
        \end{subfigure}
        \begin{subfigure}[b]{0.19\textwidth}
            \centering
            \includegraphics[width=\textwidth]{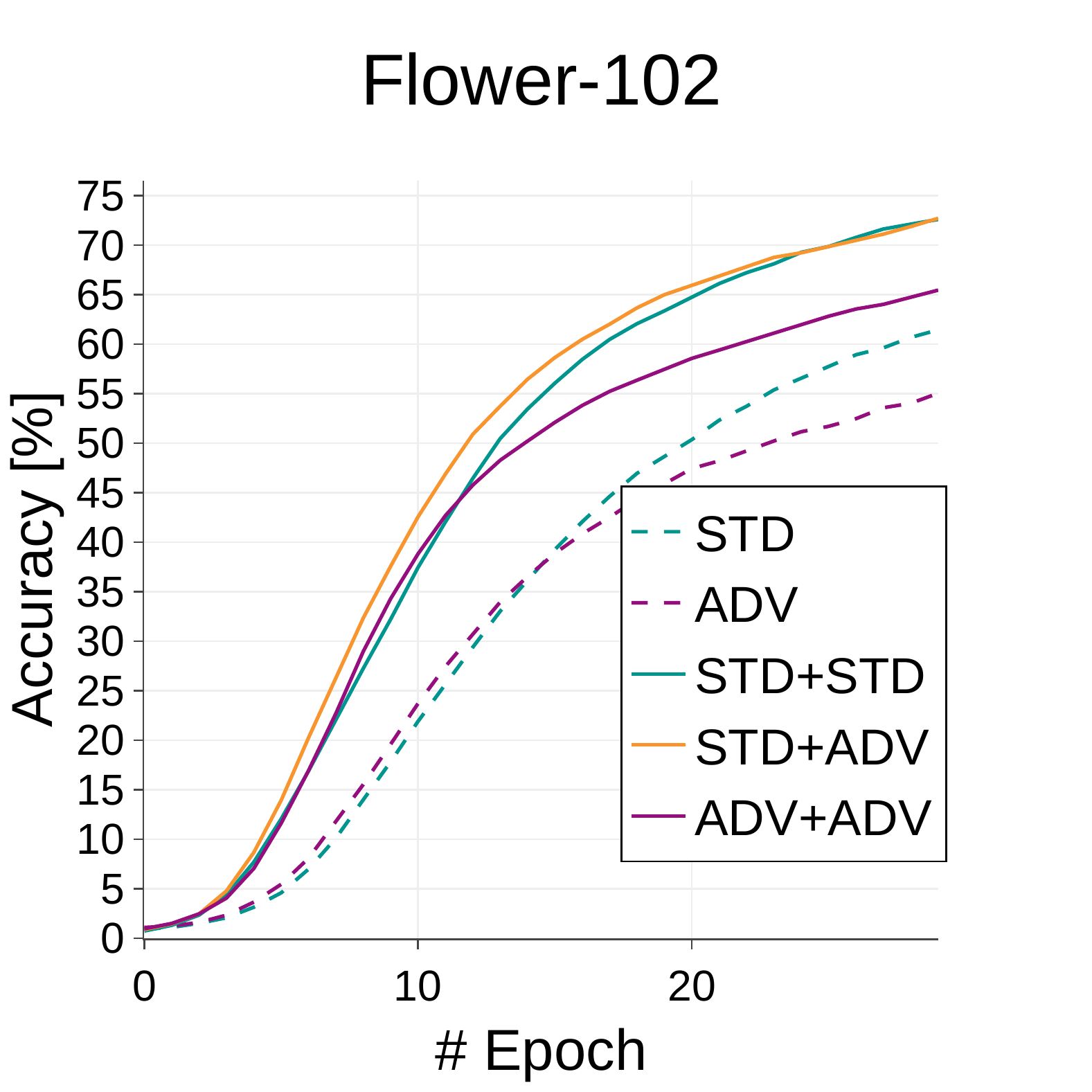}
        \end{subfigure}
        \begin{subfigure}[b]{0.19\textwidth}
            \centering
            \includegraphics[width=\textwidth]{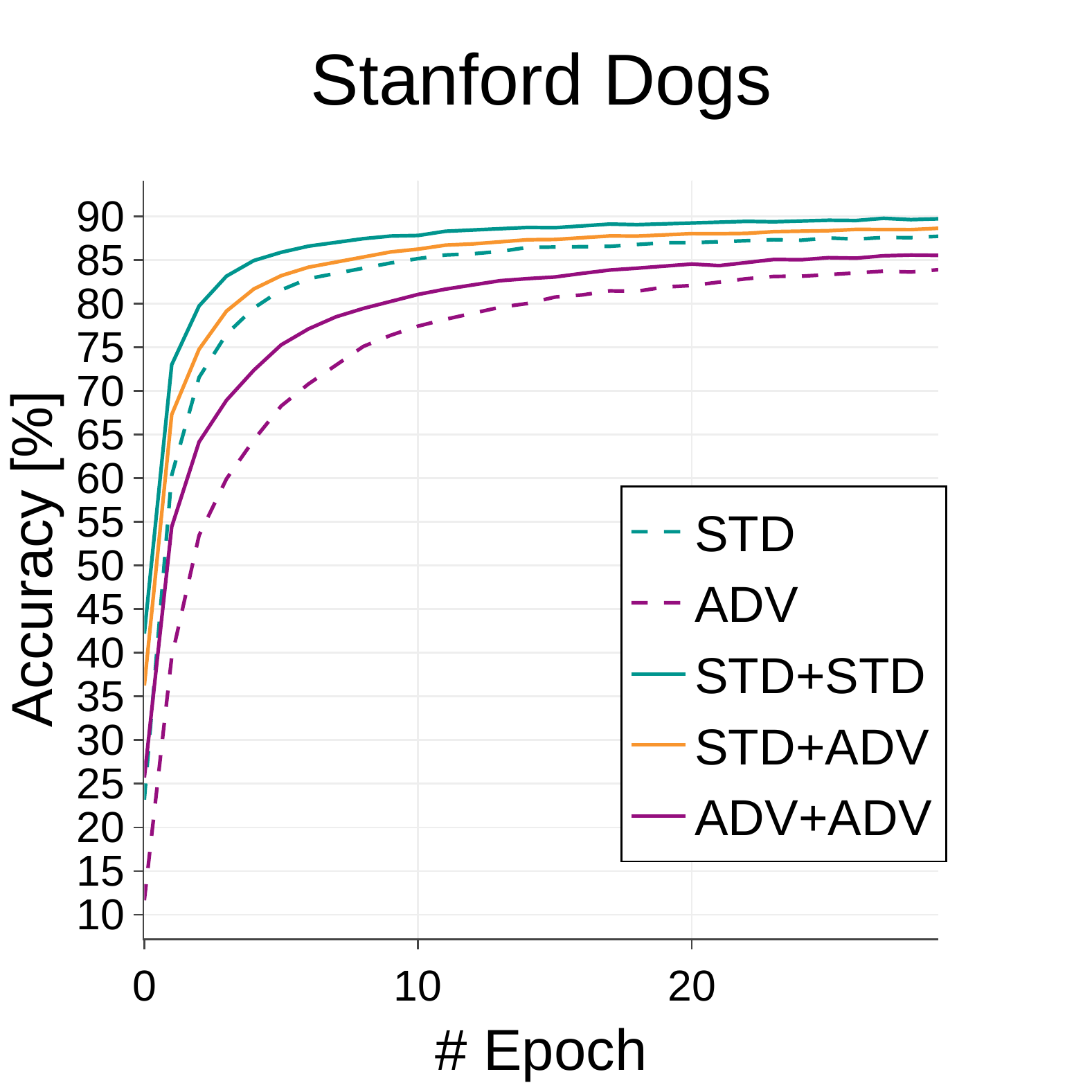}
        \end{subfigure}
        \begin{subfigure}[b]{0.19\textwidth}
            \centering
            \includegraphics[width=\textwidth]{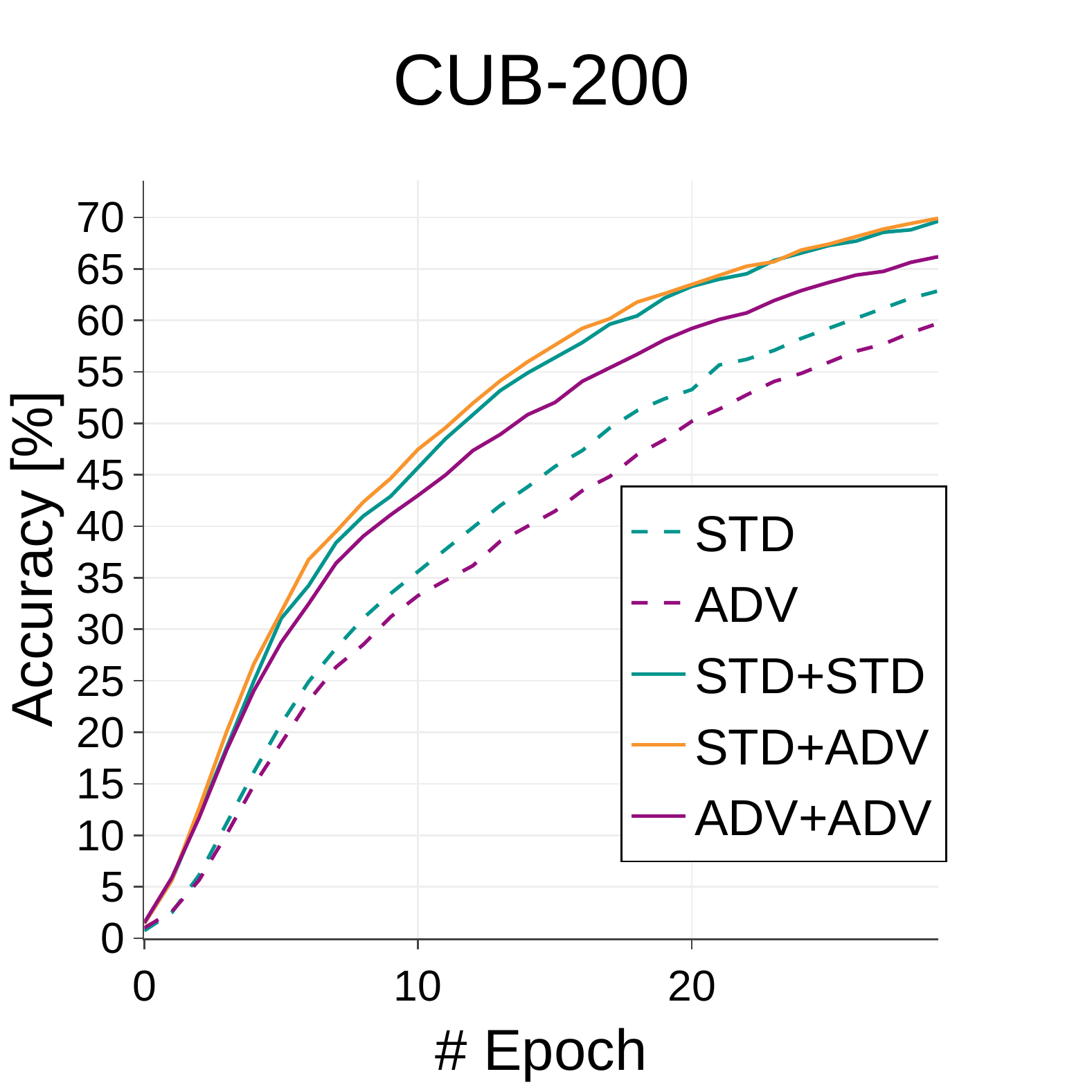}
        \end{subfigure}
        \begin{subfigure}[b]{0.19\textwidth}
            \centering
            \includegraphics[width=\textwidth]{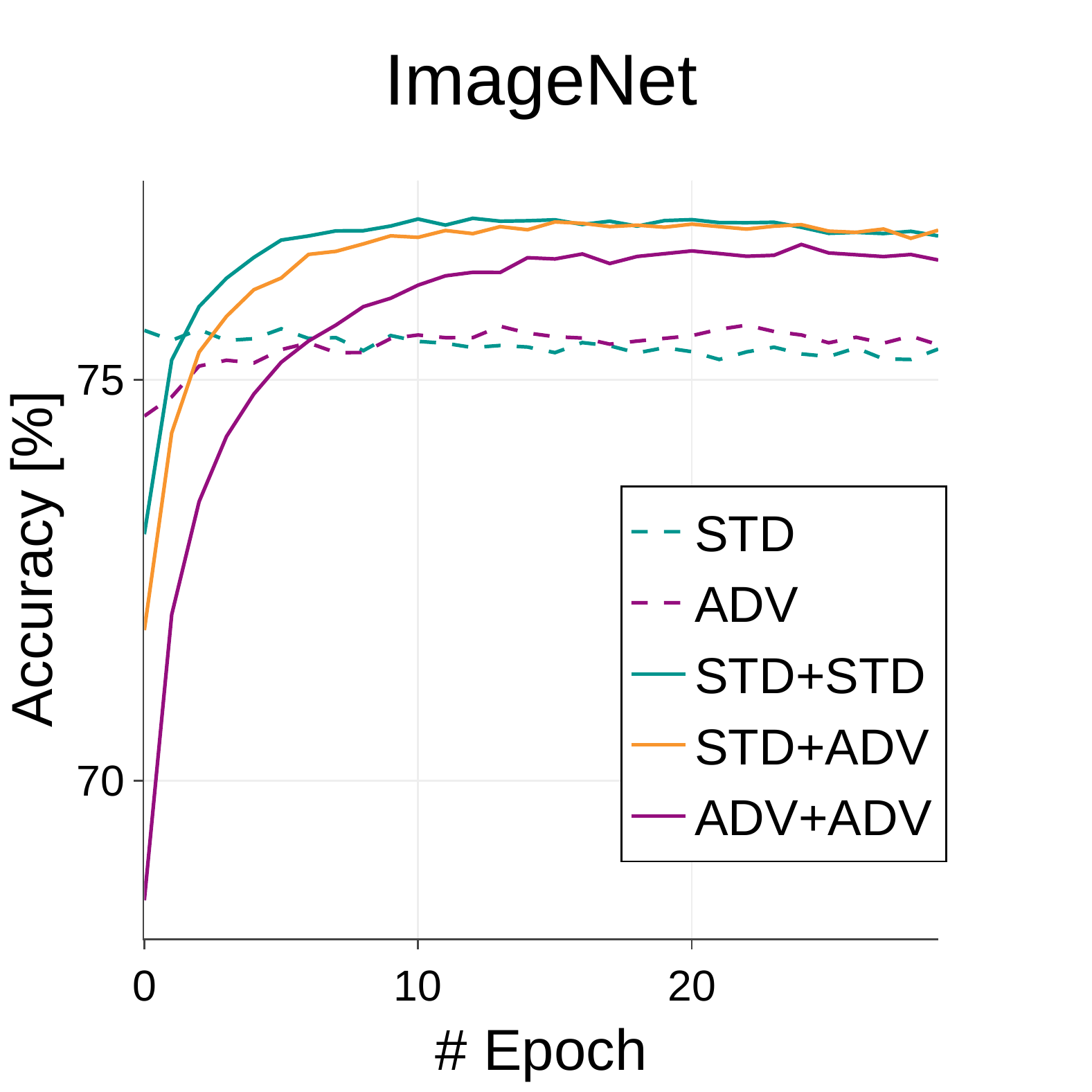}
        \end{subfigure}
        \caption{ResNet-50}
    \end{subfigure}
    \caption{Learning curves of AlexNet and ResNet-50 on 10 datasets. The solid lines mean ensemble models and the dashed lines mean models without ensemble learning.}
    \label{fig:ensemble}
\end{figure*}

Figure~\ref{fig:ensemble} shows the results of ensemble learning for AlexNet and ResNet-50.
As can be seen from the results, the STD+ADV models outperform the plain STD and plain ADV models in most datasets.
On the other hand, the STD+STD models tend to be more accurate for the high resolution datasets (e.g., STL-10, Food-101, Flower-102, Stanford Dogs, CUB-200 and ImageNet) and the ADV+ADV models for the low resolution datasets (e.g., SVHN, CIFAR-10, CIFAR-100, Tiny ImageNet)
While the the accuracy of the STD+STD and ADV+ADV models fluctuate depending on the resolution of the datasets, the STD+ADV models show the comparable accuracy to the highest-accurate models in all the datasets.
These results indicate that the STD+ADV models obtain resolution-invariant features and see the inputs at various scales.
From the above, it has been considered that the adversarial training has contradictory to the accuracy, but we clarified that both standard accurate and adversarially robust features can be used to improve the accuracy.

\section{Discussion}
Through various experiments, we confirmed that adversarially trained models pay attention to larger scale patterns and stronger edges rather than fine textures.
Here, we would like to further consider that the visualization results of adversarially trained models suggest.
On the first layer visualization in Figure~\ref{fig:first_layer}, both the standard trained and the adversarially trained models basically obtained similar patterns, but there exist two crucial differences. 
The first one is, as aforementioned in Section~\ref{subsec:what_do}, the values of adversarially trained ones are smaller than standard trained ones.
The other noticeable difference is that grid-like or checkerboard-like patterns appear in the adversarially trained models (see the enlarged images placed at the bottom left corners in Figure \ref{fig:first_layer}).
The shape of the patterns appear in ResNet-50 and ResNet-101 looks similar although that of AlexNet is different.
These observations might suggest that to some extent these patterns depend on the model architecture and become a clue to explain why adversarial examples are well transferred in the same architecture family as reported in \cite{su2018robustness}.
As future works, we will analyze this result in detail and apply it to construct more effective defense methods or strong perturbations which have high transferability.

\section{Conclusion}
In this paper, we address an open question: ``What do adversarially robust models look at?'' and provided the visual and the quantitative evidence that adversarially robust models looks at things at a larger scale than standard accurate models.
Furthermore, we showed empirically that both standard trained and adversarially trained features are useful for improving accuracy contrary to the previous reports that adversarial robustness decreases the accuracy.
From this result, we believe that adversarially robust features may also be useful in improving the accuracy of tasks other than classification (e.g., surface normal estimation and keypoint detection) and analyzing them may give clues not only to devise effective defense methods against adversarial perturbations but also to clarify the blackbox mechanisms of DNNs.
Like these, exploring the potential of adversarially robust features is our future work.

{\small
\bibliographystyle{ieee}
\bibliography{egbib}
}

\clearpage

\setcounter{section}{0}
\renewcommand{\thesection}{\Alph{section}}

\section{Sanity Checks of Model Parameters Dependence}
In the main paper, we used some visualization methods to confirm visually the difference between feature representations of standard trained models and those of adversarially trained ones.
While visualization is one of the very important ways to explain the predictions of DNNs, some recent works claimed that a part of them lacks the theoretical background and does not explain the network decisions correctly.

In particular, guided backpropagation~\cite{springenberg2015striving}, one of the visualization methods we used, has been shown to lack class-discriminability~\cite{zhou2016learning,selvaraju2017grad,nie2018atheoretical}.
However, the lack of this class dependence does not matter in achieving our goal of clarifying the difference between what the models look at.
On the other hand, Adebayo et al.~\cite{adebayo2018sanity}\footnote{Their GitHub page (\url{https://github.com/adebayoj/sanity_checks_saliency}) exists, but it has no implementations.} pointed out the problem that sensitivity maps by guided backpropagation are independent on the model parameters, and this problem is critical for our goal.
Thus, in this section, we conduct the sanity check of this model parameters independence for various visualization methods and experimental results show that guided backpropagation has the model parameters dependence unlike their results~\cite{adebayo2018sanity}.
We release our code\footnote{\url{https://github.com/anonymous-author-iccv2019/sanity_checks}} for reproducibility of our experiments.

Besides this, other desirable characteristics as the attribution methods have been discussed in many works~\cite{zhou2016learning,kindermans2017unreliability,ghorbani2017interpretation,selvaraju2017grad,sundararajan2017axiomatic,nie2018atheoretical}.
Note that we agree their view that reliance solely on visual assessment can be misleading, and also confirmed that our claim is correct via several quantitative experiments in the main paper.

\subsection{Mathematical Formulation}
In this section, we introduce several visualization methods that we use in our sanity checks.
There are some parts that overlap with the main paper, but they are written again to make this supplementary material self-contained.
The element-wise product of the input and the gradient (converted to gray-scale) is often used to reduce visual illegibility, but we use the raw gradients here because the color information of the gradients can be also important.
Therefore we omit the element-wise product in the formulations accordingly.

Given a $K$-class classifier $f: \mathbb{R}^d \to \mathbb{R}^K$ and a pair of a data point $\boldsymbol{x} \in \mathbb{R}^d$ and a ground-truth label $t$, the predicted label is obtained by $\hat{k}(\boldsymbol{x})= {\rm arg~max}_{k} f_k (\boldsymbol{x})$, where $f_k(\boldsymbol{x})$ is the $k$-th component of $f(\boldsymbol{x})$ that corresponds to the $k$-th class.

\noindent
{\bf Vanilla Gradient.}
Vanilla gradient calculates the gradient of the classifier output of the ground-truth label with respect to the input as follows:
\begin{equation}
    \boldsymbol{S}_{\rm vanilla} (\boldsymbol{x}) = \nabla_{\boldsymbol{x}} f_t (\boldsymbol{x}).
\end{equation}

\noindent
{\bf Loss Gradient.}
Loss gradient calculates the gradient of the loss with respect to the input as follows:
\begin{equation}
    \boldsymbol{S}_{\rm loss} (\boldsymbol{x}) = \nabla_{\boldsymbol{x}} L(\hat{k}(\boldsymbol{x}), t).
\end{equation}

\noindent
{\bf Guided Backpropagation.}~\cite{springenberg2015striving}
Guided backpropagation set all the negative gradients to 0 like the ReLU function.
We consider the $i$-th ReLU activation in the $l$-th layer with its input $y_i^{(l)}$ and its output $o_i^{(l)}$, and denote $\sigma(t) = \max(t, 0)$ the ReLU activation.
In the gradient calculation, the forward ReLU is formally defined as follows:
\begin{equation}
    \sigma_{f,i}^{(l)} (t) = \mathbb{I} (y_{i}^{(l)}) t
\end{equation}
where $\mathbb{I}(\cdot)$ is the indicator function, and the backward ReLU is formally defined as follows:
\begin{equation}
    \sigma_{b,i}^{(l)} (t) = \mathbb{I} (R_i^{(l)}) t
\end{equation}
where $R_i^{(l)}$ is the top gradient before activation, i.e., gradient of the output score with respect to $o_i^{(l)}$.
Then, the (modified) gradient after activation $T_i^{(l)}$, i.e., gradient of the output score with respect to $y_i^{(l)}$ is given as follows:
\begin{equation}
    T_i^{(l)} = 
    \begin{cases}
        \sigma_{f,i}^{(l)} (R_i^{(l)}) & {\rm (vanilla \; gradient)} \\
        \sigma_{f,i}^{(l)} (\sigma_{b,i}^{(l)} (R_i^{(l)})) & {\rm (guided \; backpropagation)}.
    \end{cases}
\end{equation}

\noindent
{\bf SmoothGrad.}~\cite{smilkov2017smoothgrad}
In order to avoid noisy results, SmoothGrad average the gradients with respect to the multiple noise added input as follows:
\begin{equation}
    \boldsymbol{S}_{\rm smooth} (\boldsymbol{x}) = \frac{1}{N} \sum_{n=1}^{N} \nabla_{\boldsymbol{x}} f_t (\boldsymbol{x} + \mathcal{N}(0, \sigma^2))
\end{equation}
where $N$ is the number of samples and $\mathcal{N}(0, \sigma^2)$ represents Gaussian noise with standard deviation $\sigma$.

\noindent
{\bf Integrated Gradient.}~\cite{sundararajan2017axiomatic}
Integrated gradient accumulates the gradients at all points along the path between the input $\boldsymbol{x}$ and the baseline $\boldsymbol{x}'$ as follows:
\begin{equation}
    \boldsymbol{S}_{\rm integrated} (\boldsymbol{x}) = \int_{\alpha=0}^{1} \nabla_{\boldsymbol{x}} f_t (\boldsymbol{x}' + \alpha (\boldsymbol{x} - \boldsymbol{x}')) d \alpha . 
\end{equation}

\subsection{Experimental Results and Discussion}
\begin{figure*}[h!]
    \centering
    \includegraphics[height=0.99\textheight]{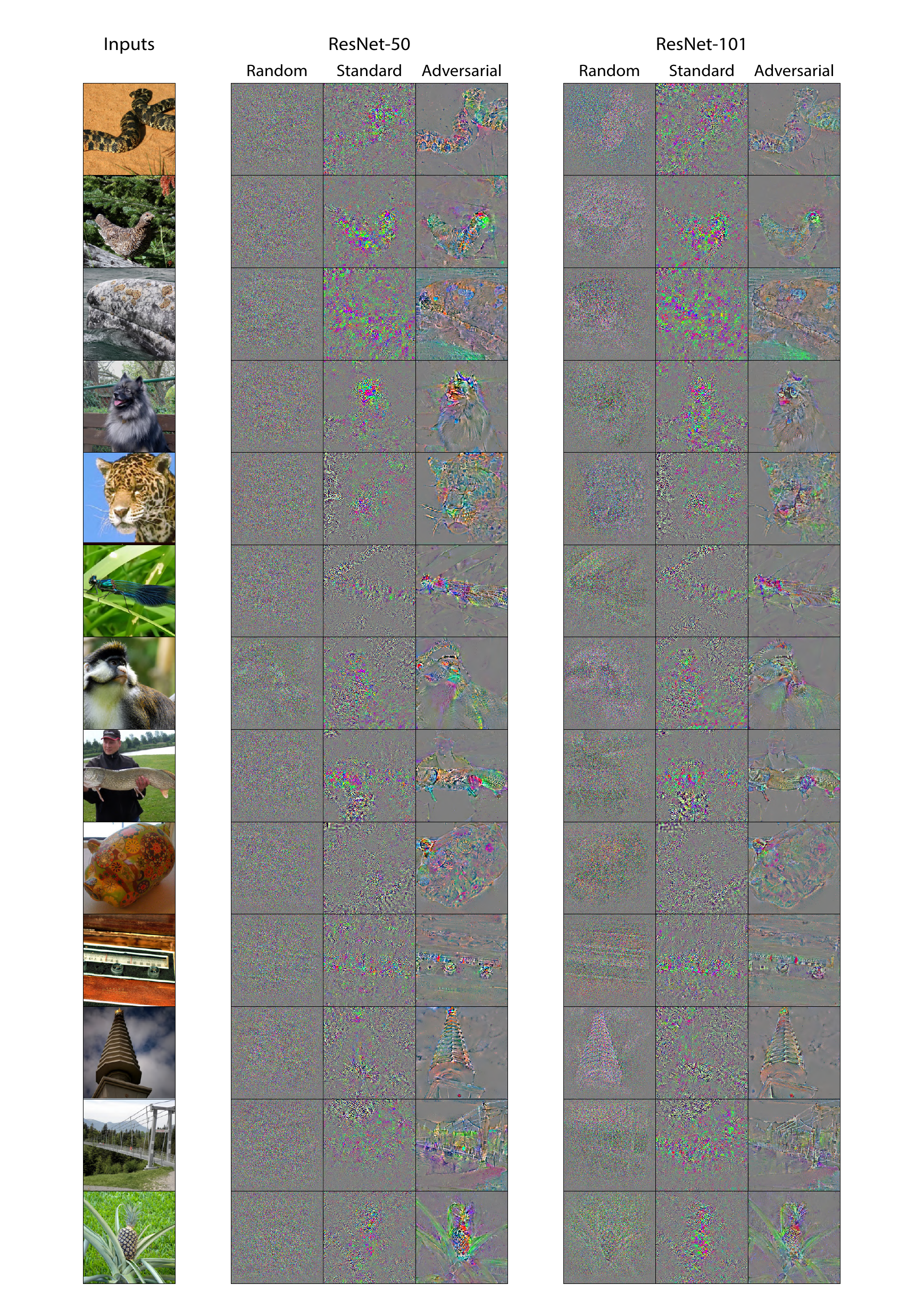}
    \caption{Visualization results of the sensitivity maps by Vanilla Gradient.}
    \label{fig:map_supp_vanilla}
\end{figure*}

\begin{figure*}[h!]
    \centering
    \includegraphics[height=0.99\textheight]{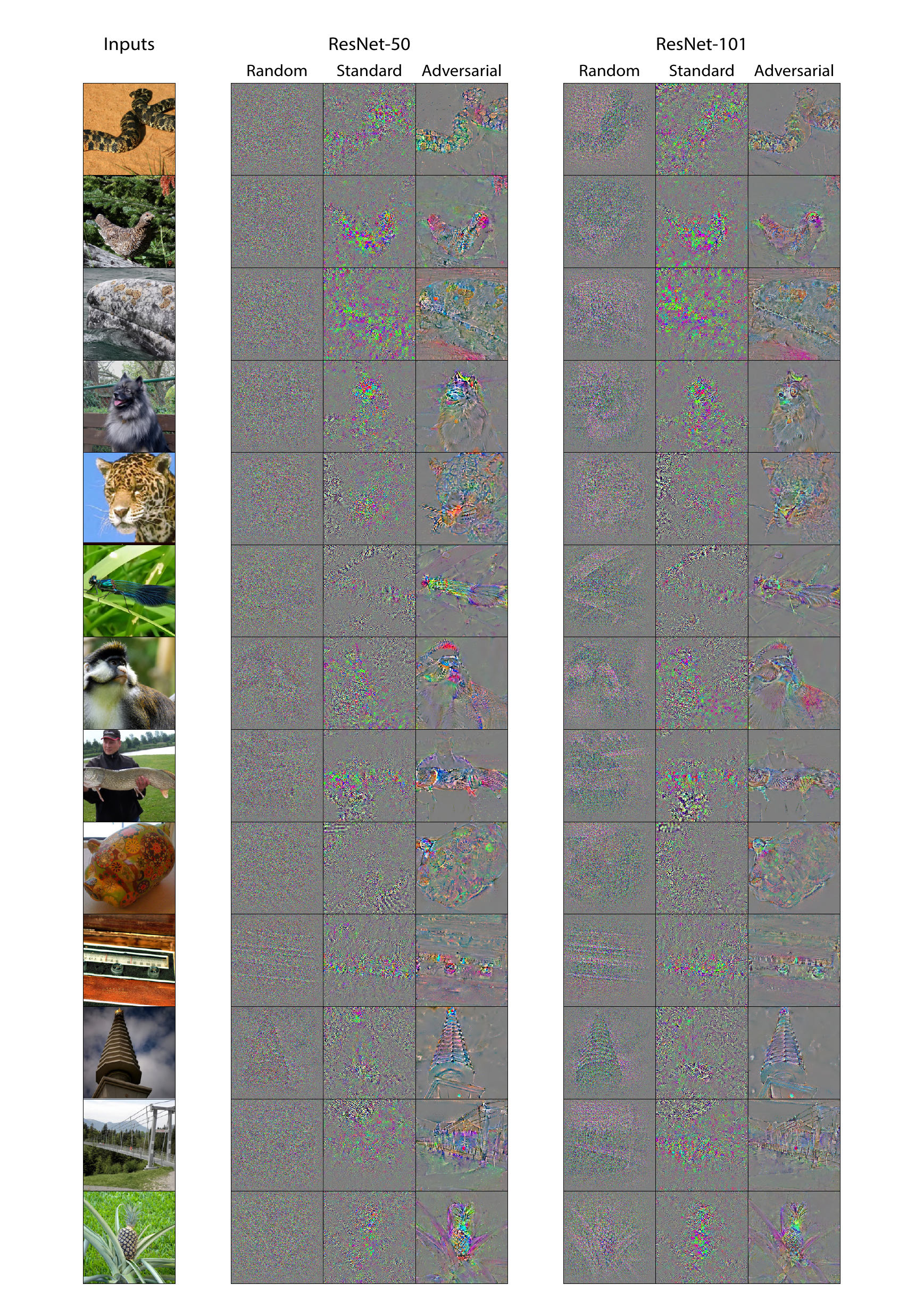}
    \caption{Visualization results of the sensitivity maps by Loss Gradient.}
    \label{fig:map_supp_loss}
\end{figure*}

\begin{figure*}[h!]
    \centering
    \includegraphics[height=0.99\textheight]{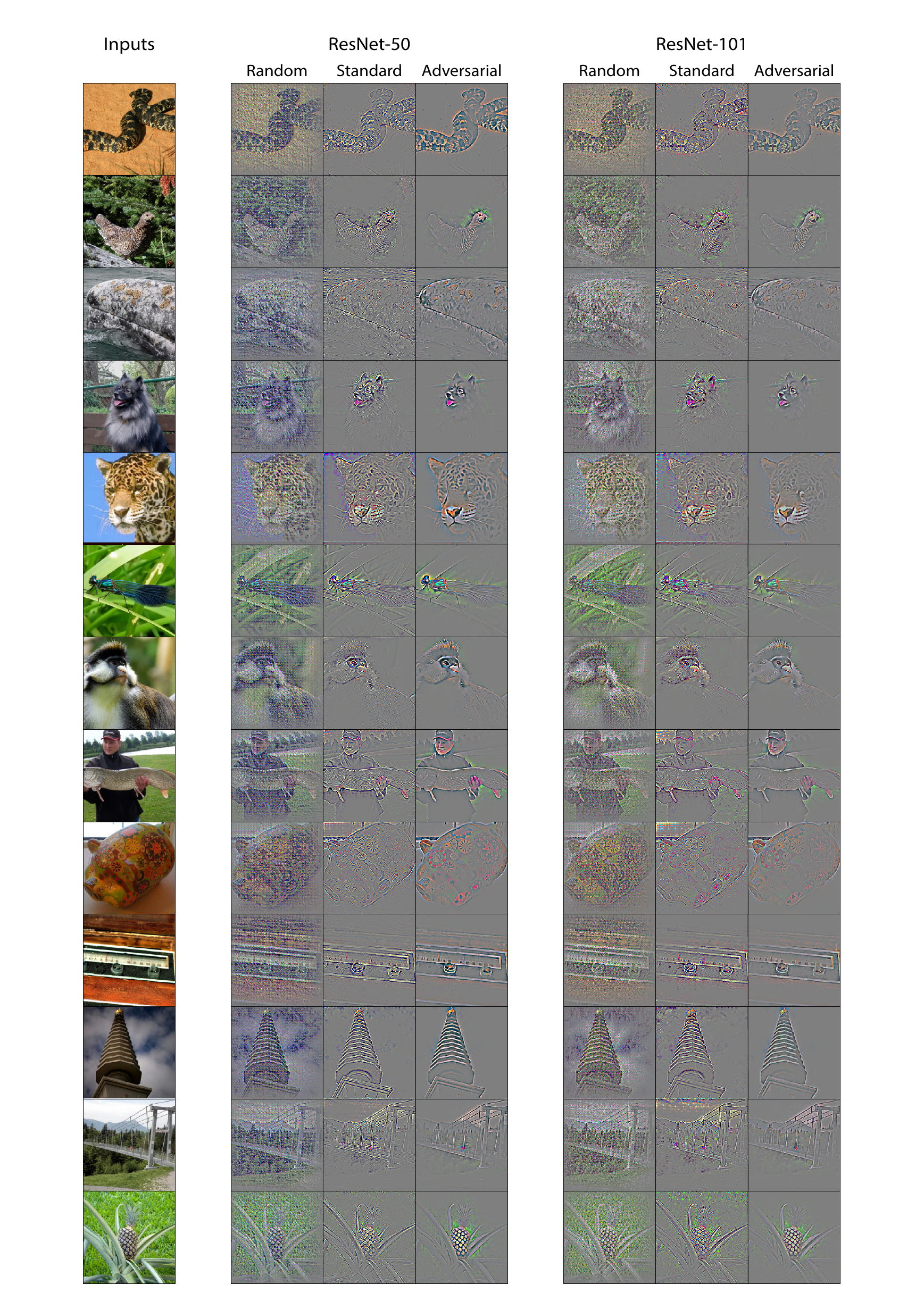}
    \caption{Visualization results of the sensitivity maps by Guided Backpropagation.}
    \label{fig:map_supp_guided}
\end{figure*}

\begin{figure*}[h!]
    \centering
    \includegraphics[height=0.99\textheight]{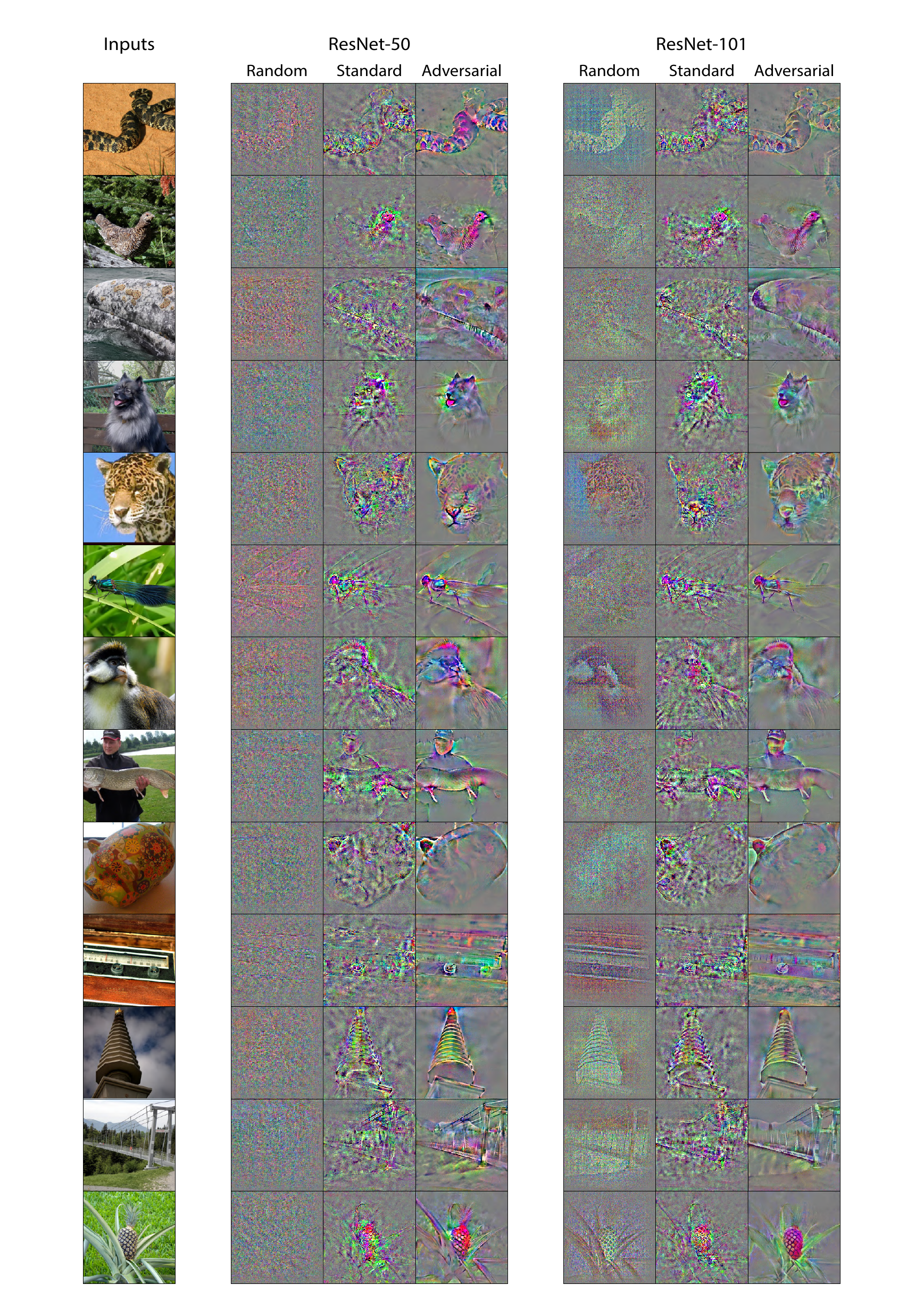}
    \caption{Visualization results of the sensitivity maps by SmoothGrad.}
    \label{fig:map_supp_smooth}
\end{figure*}

\begin{figure*}[h!]
    \centering
    \includegraphics[height=0.99\textheight]{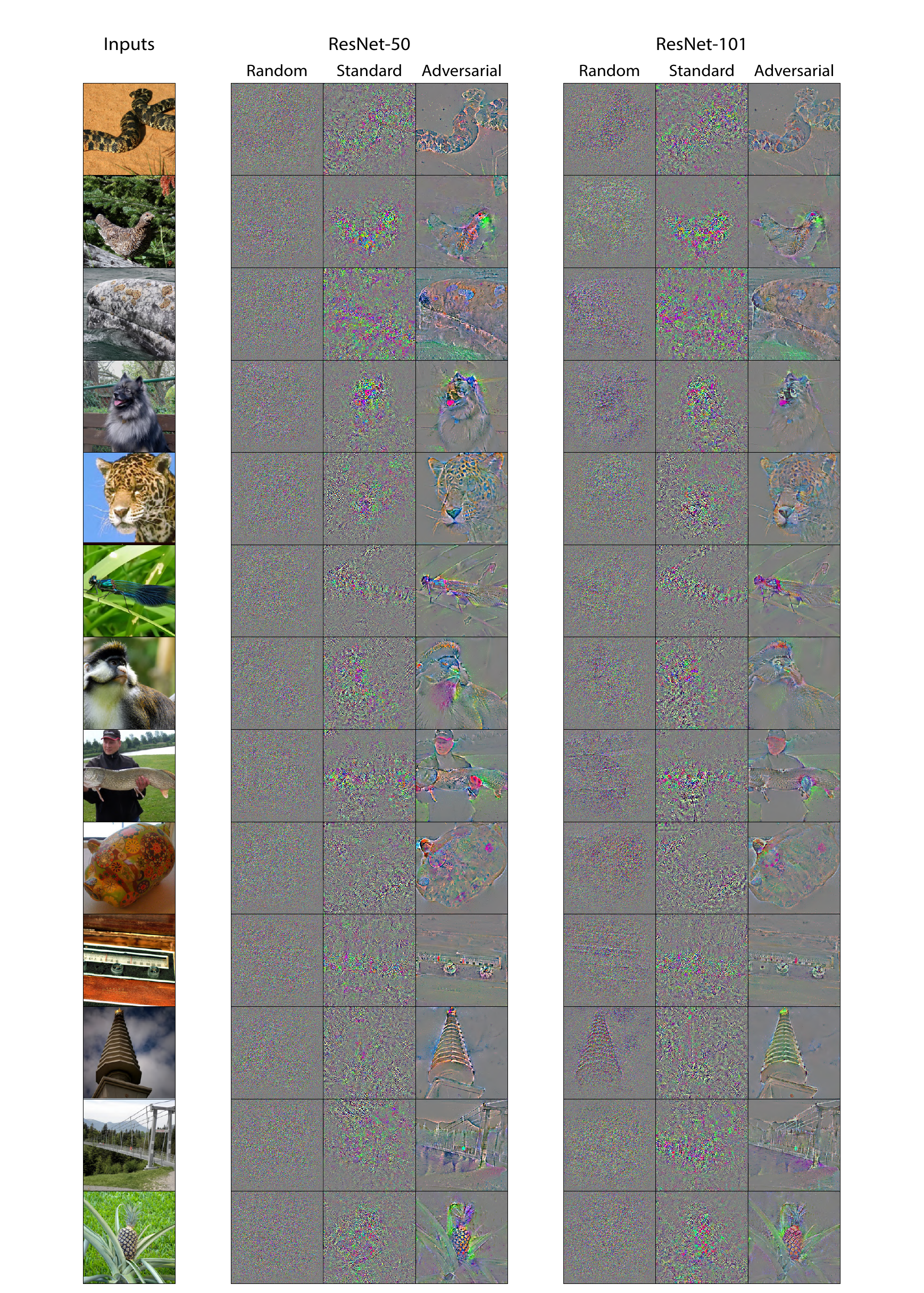}
    \caption{Visualization results of the sensitivity maps by Integrated Gradient.}
    \label{fig:map_supp_integrated}
\end{figure*}

We performed aforementioned visualization methods on randomly initialized, standard trained and adversarially trained models for ResNet-50 and ResNet-101 as shown in
Figure~\ref{fig:map_supp_vanilla} - \ref{fig:map_supp_integrated}.
As it can be seen in Figure~\ref{fig:map_supp_guided}, guided backpropagation leads to very different visualization results when the model parameters are different, similar to other visualization methods.
Therefore, it can be concluded that guided backpropagation has ``model parameters dependence'' and this phenomenon has been also confirmed in the previous report~\cite{nie2018atheoretical}.

In addition, we can see that guided backpropagation is the best suited to determine whether the models look at fine textures or not.
This can be thought to be due to the strong ability of guide backpropagation to generate clearer outputs.
Nie et al.~\cite{nie2018atheoretical} proved theoretically that guided backpropagation recovers the input images on randomly initialized models, and highlight more relevant pixels when the model parameters are strongly biased via training, as can be confirmed in Figure~\ref{fig:map_supp_guided}.
We believe that it is necessary to compare with an appropriate baseline, rather than judge them independently, especially when using powerful visualization methods like guided backpropagation.
The need for a baseline is also stated in the existing work~\cite{sundararajan2017axiomatic}.
In fact, comparing the results in models with different parameters led to helpful clues to assess their characteristics.

Some people may think it strange that the results on randomly initialized models are human-interpretable in almost visualization methods.
This phenomenon can be thought to be rooted in the local connectivity of CNNs.
Noroozi et al.~\cite{noroozi2016unsupervised} reported that AlexNet can achieve 12\% accuracy in ImageNet (the chance rate is 0.01\%) by training only the fully connected layers even though the parameters of all convolutional layers are fixed with random initialization (from a Gaussian distribution).
Based on this strong prior, Caron et al.~\cite{caron2018deep} proposed an unsupervised learning method by clustering iteratively deep features and using these cluster assignments as pseudo-labels to learn the parameters of the convolution part.
Therefore, this phenomenon is not the result of the defects of visualization methods and is due to the potential of CNNs.


\section{Additional Sensitivity Map Visualization}
\begin{figure*}[h!]
    \centering
    \includegraphics[height=0.99\textheight]{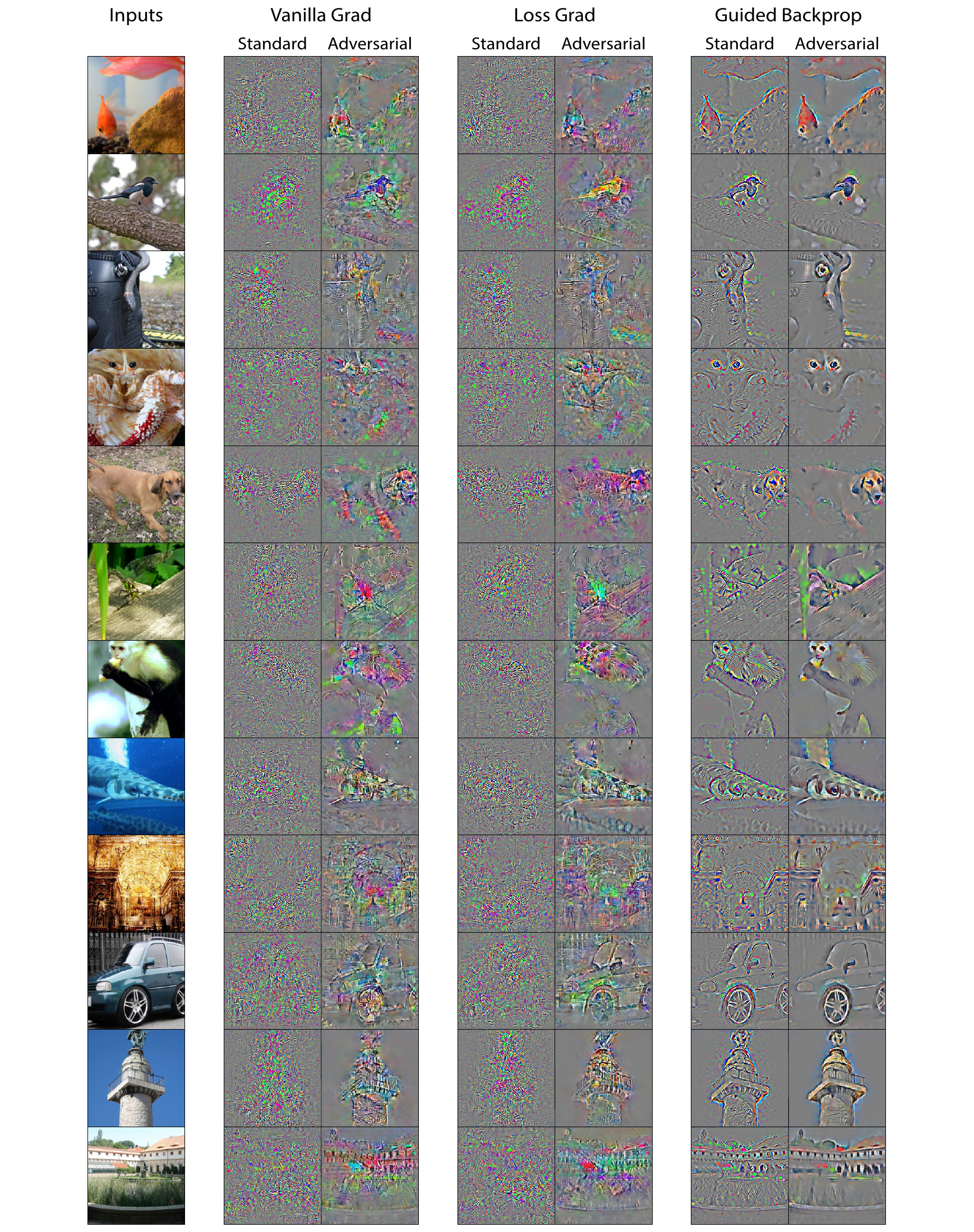}
    \caption{Additional visualization results of the sensitivity maps of AlexNet.}
    \label{fig:map_supp_alexnet}
\end{figure*}
\begin{figure*}[h!]
    \centering
    \includegraphics[height=0.99\textheight]{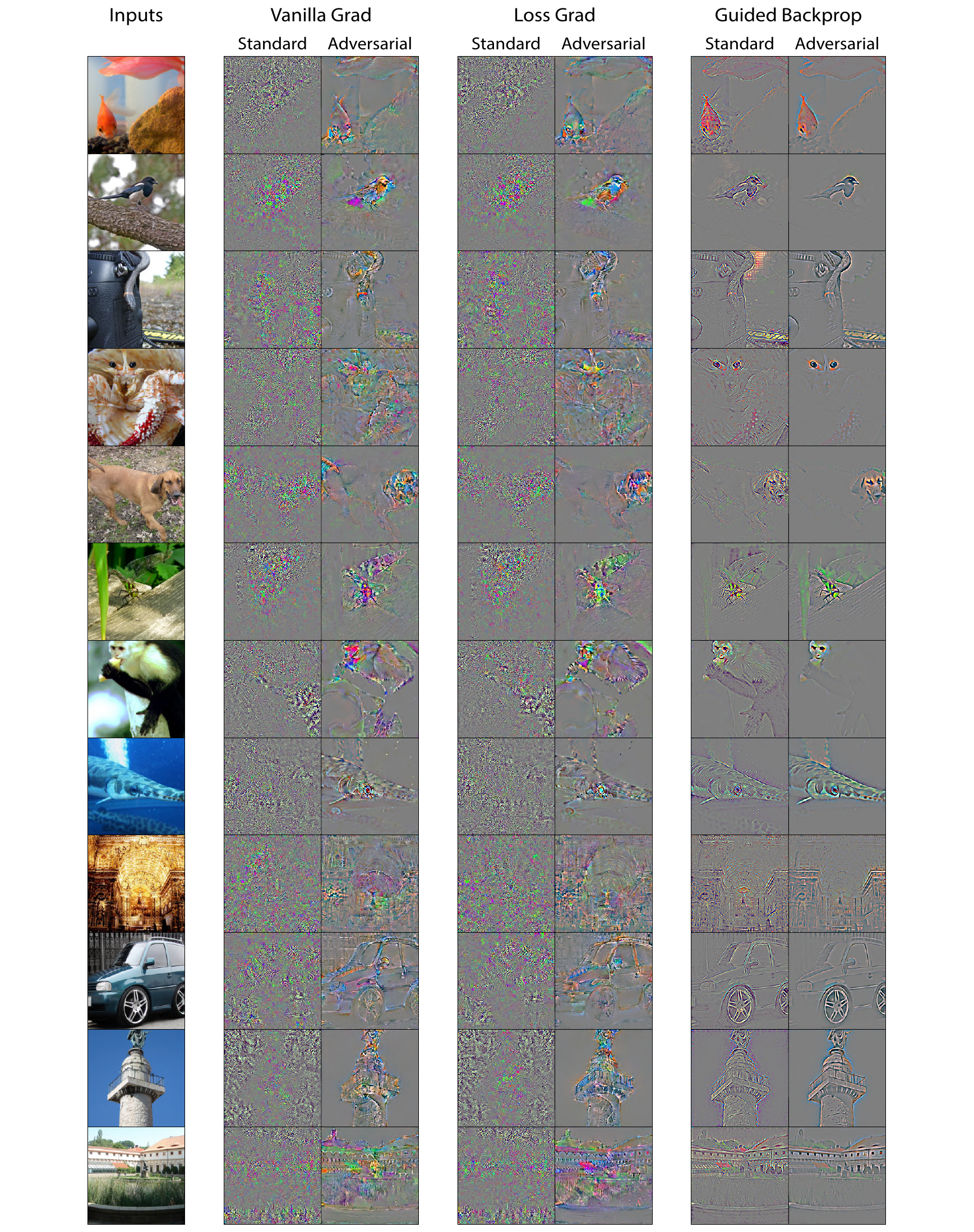}
    \caption{Additional visualization results of the sensitivity maps of ResNet-50.}
    \label{fig:map_supp_resnet50}
\end{figure*}

We provide more visualization results of the sensitivity maps of standard trained and adversarially trained networks (AlexNet and ResNet-50).
The input images are randomly chosen from the ImageNet dataset. As we can see, all the results (Figure~\ref{fig:map_supp_alexnet} and Figure~\ref{fig:map_supp_resnet50}) are consistent with our previous empirical observations that adversarially trained models look at things at a larger scale than standard models and pay less attention to fine textures.

\section{Additional Deeper Layer Visualization}
\begin{figure*}[t]
    \centering
    \includegraphics[width=\textwidth]{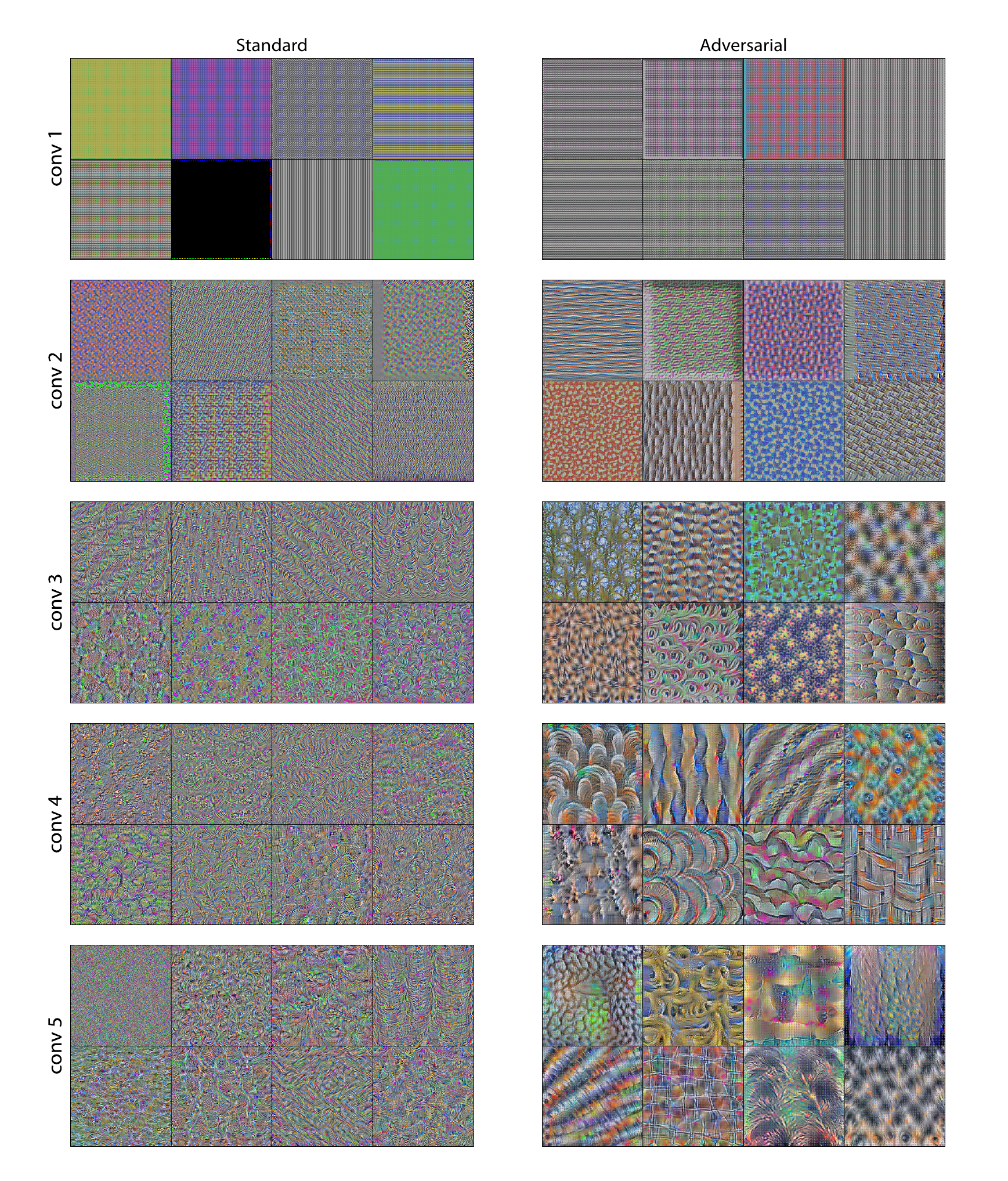}
    \caption{Deeper layer visualization of AlexNet.}
    \label{fig:dark_arts_supp_alexnet}
\end{figure*}

\begin{figure*}[h]
    \centering
    \includegraphics[width=\textwidth]{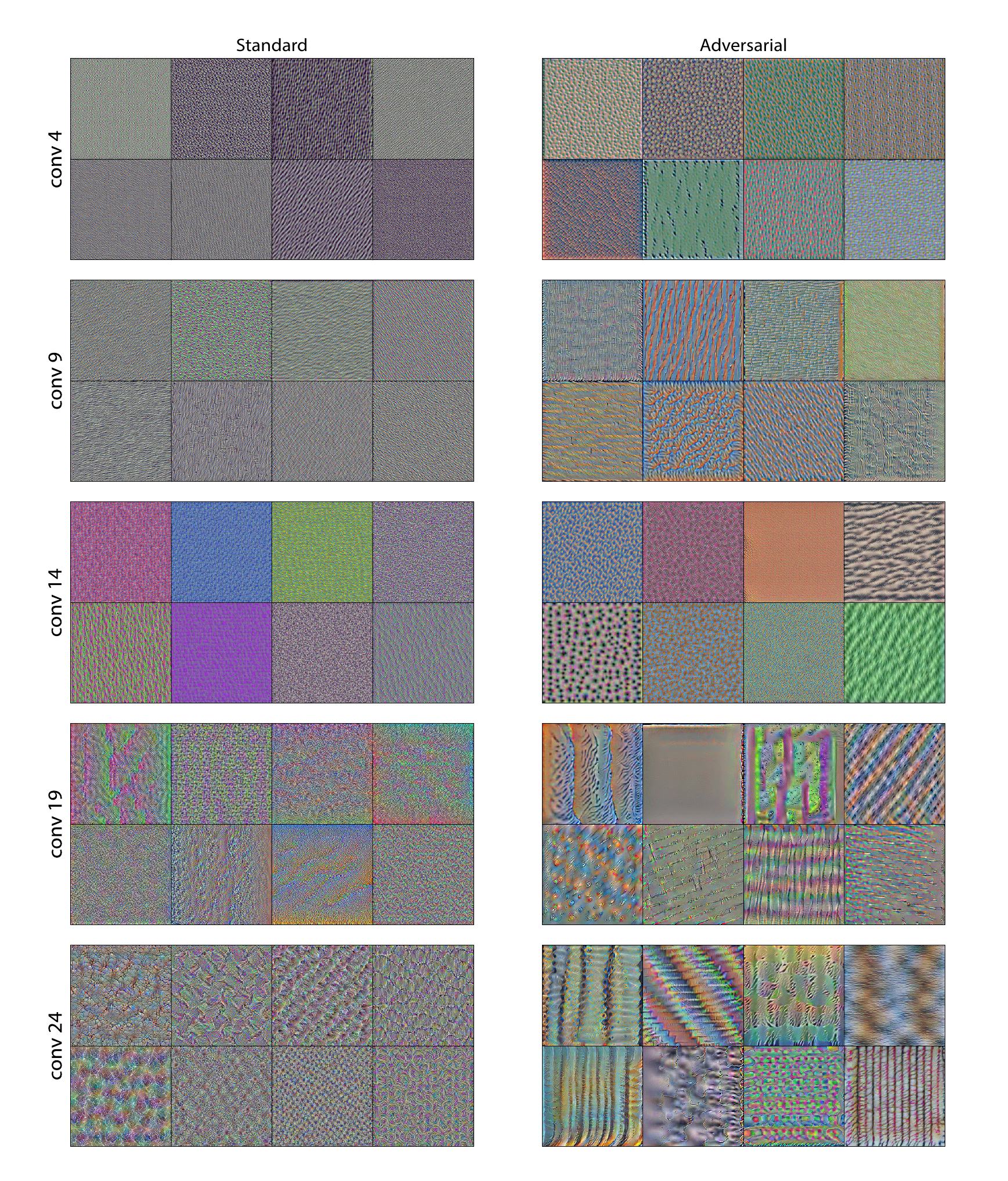}
    \caption{Additional deeper layer visualization of ResNet-50 (from conv 9 to 24).}
    \label{fig:dark_arts_supp_resnet50_formar}
\end{figure*}
\begin{figure*}[h]
    \centering
    \includegraphics[width=\textwidth]{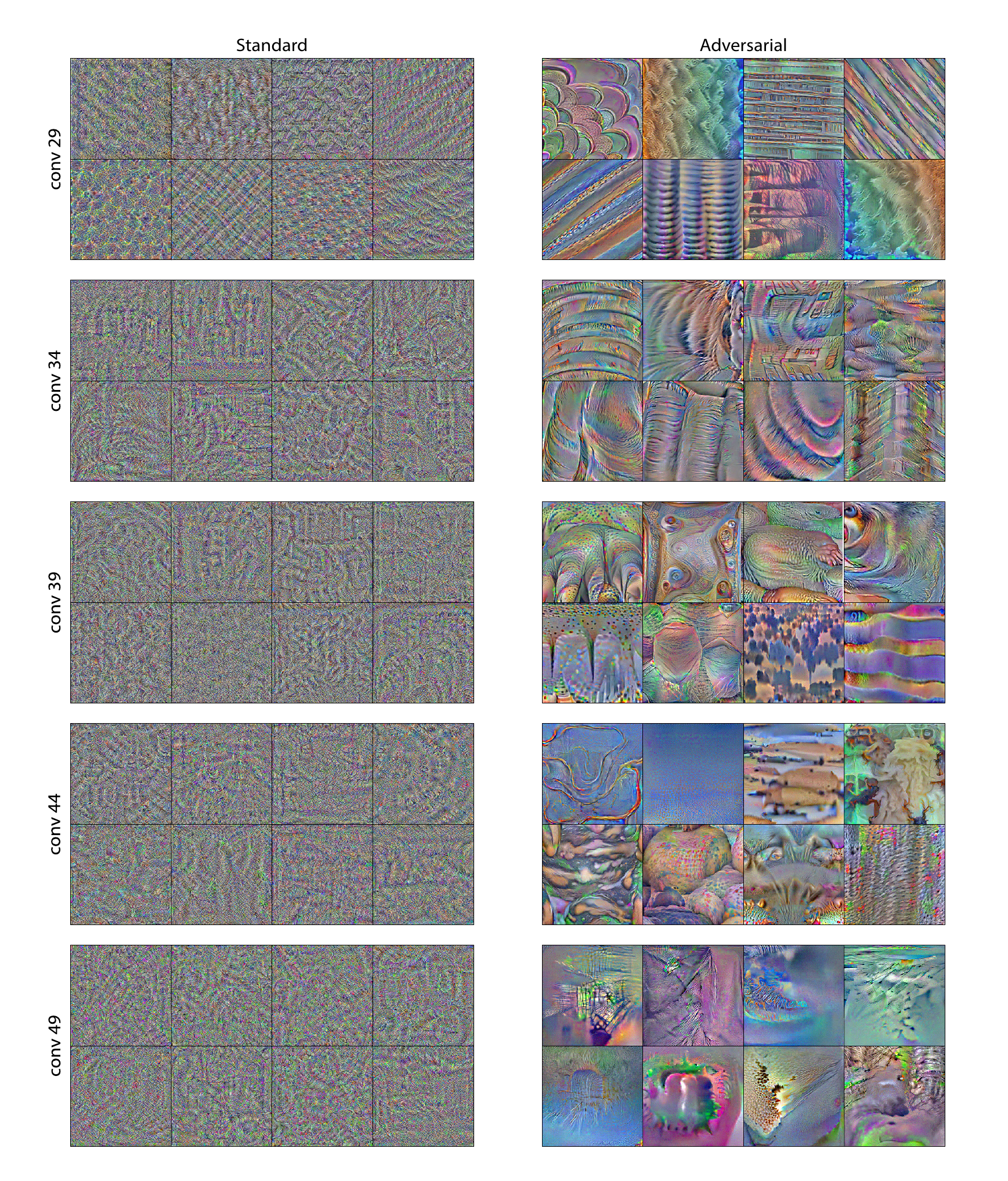}
    \caption{Additional deeper layer visualization of ResNet-50 (from conv 29 to 49).}
    \label{fig:dark_arts_supp_resnet50_latter}
\end{figure*}

\begin{figure*}[h]
    \centering
    \includegraphics[width=\textwidth]{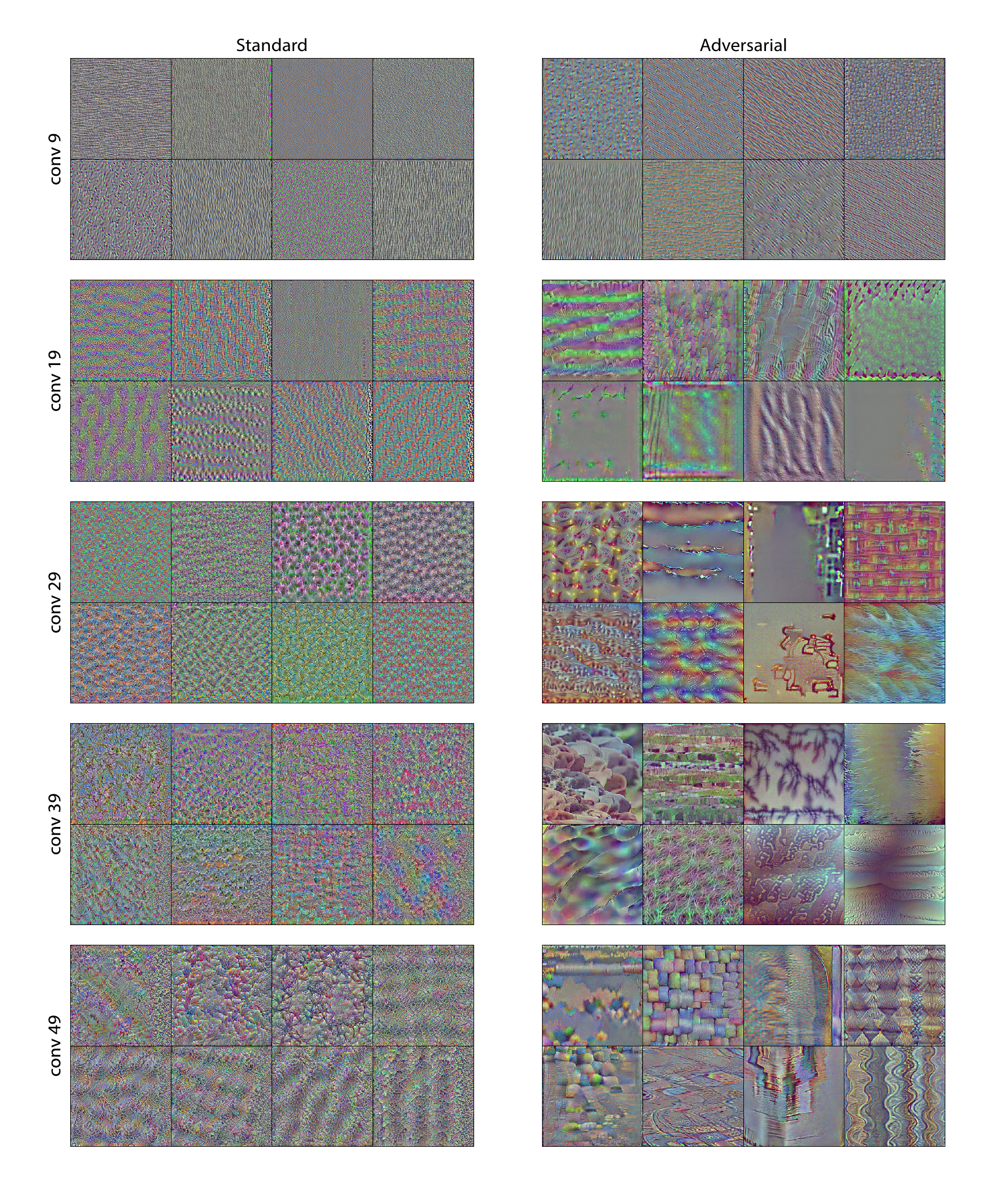}
    \caption{Additional deeper layer visualization of ResNet-101 (from conv 9 to 49).}
    \label{fig:dark_arts_supp_resnet101_formar}
\end{figure*}
\begin{figure*}[h]
    \centering
    \includegraphics[width=\textwidth]{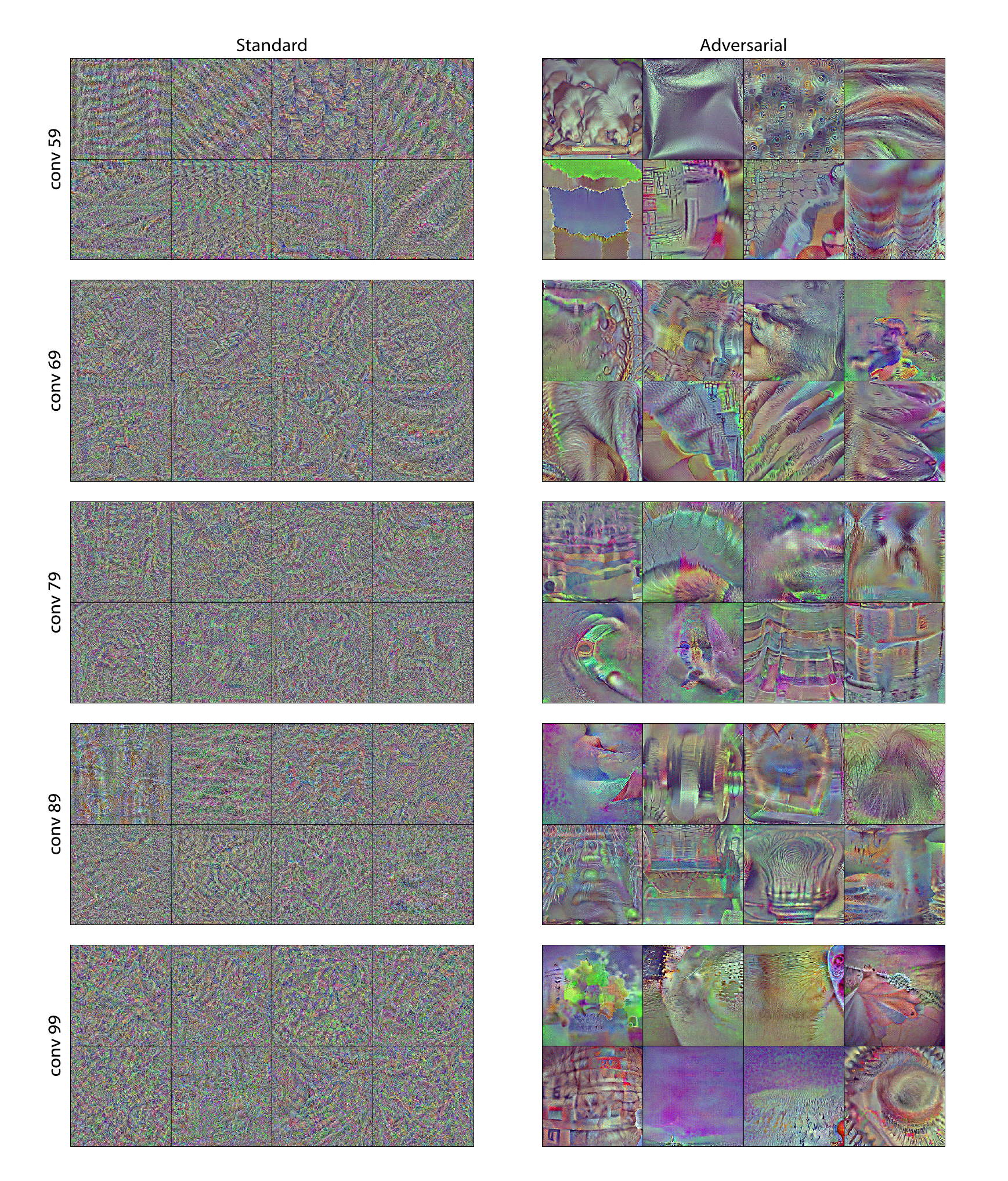}
    \caption{Additional deeper layer visualization of ResNet-101 (from conv 59 to 99).}
    \label{fig:dark_arts_supp_resnet101_latter}
\end{figure*}

In this section, we provide additional deeper layer visualizations~\cite{erhan2009visualizing} of standard trained and adversarially trained models (Figures ~\ref{fig:dark_arts_supp_alexnet}~-~\ref{fig:dark_arts_supp_resnet101_latter}).
All results are randomly selected from the indicated convolutional layers. As we can see, deeper layers of adversarially trained models capture obviously larger structures than the standard trained models.

\section{Example Images of Variants on ImageNet}
\begin{figure*}[h]
    \centering
    \includegraphics[width=\textwidth]{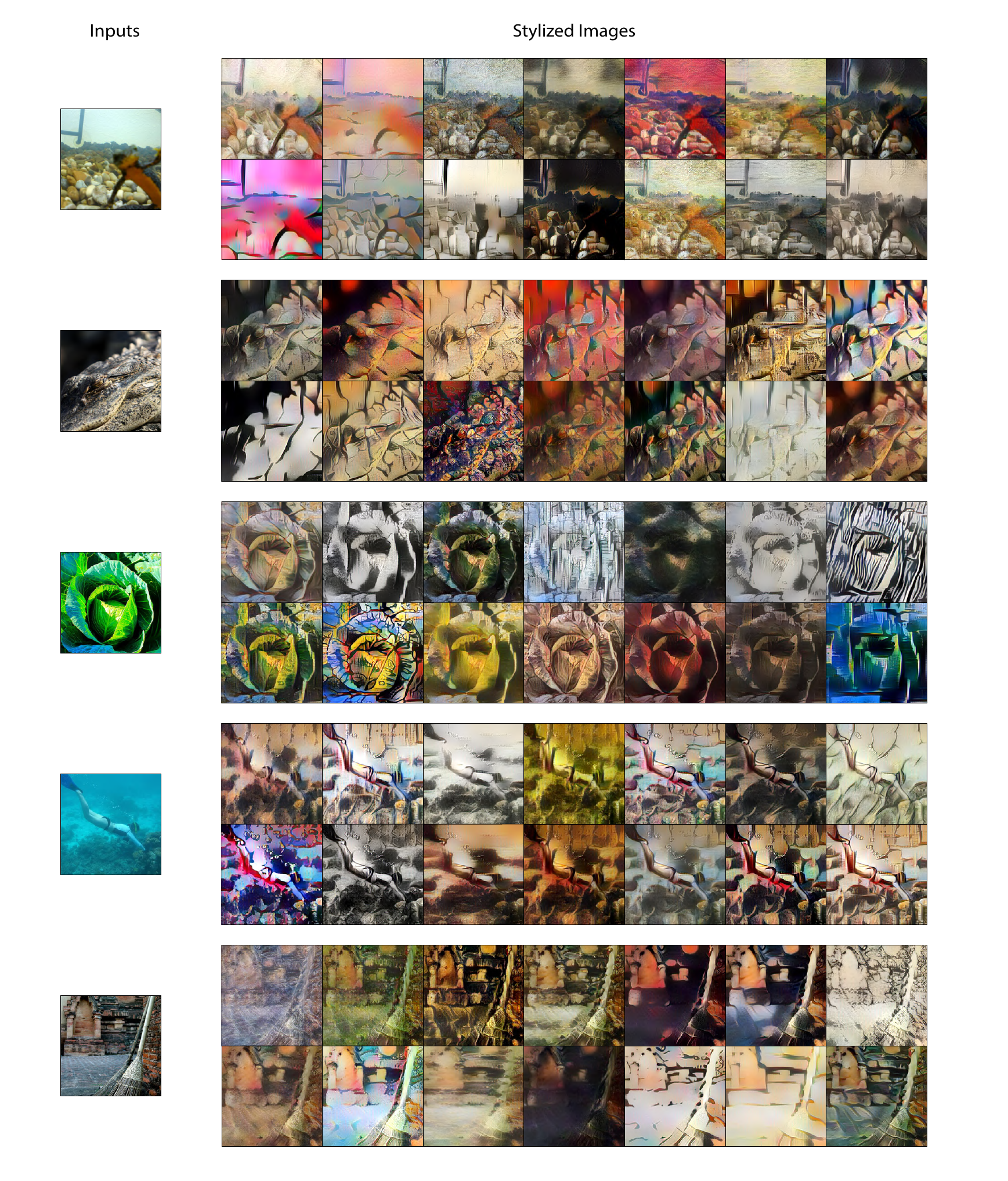}
    \caption{Example images of Stylized ImageNet. We randomly selected 14 different styles for each input image.}
    \label{fig:sin_supp_formar}
\end{figure*}
\begin{figure*}[h]
    \centering
    \includegraphics[width=\textwidth]{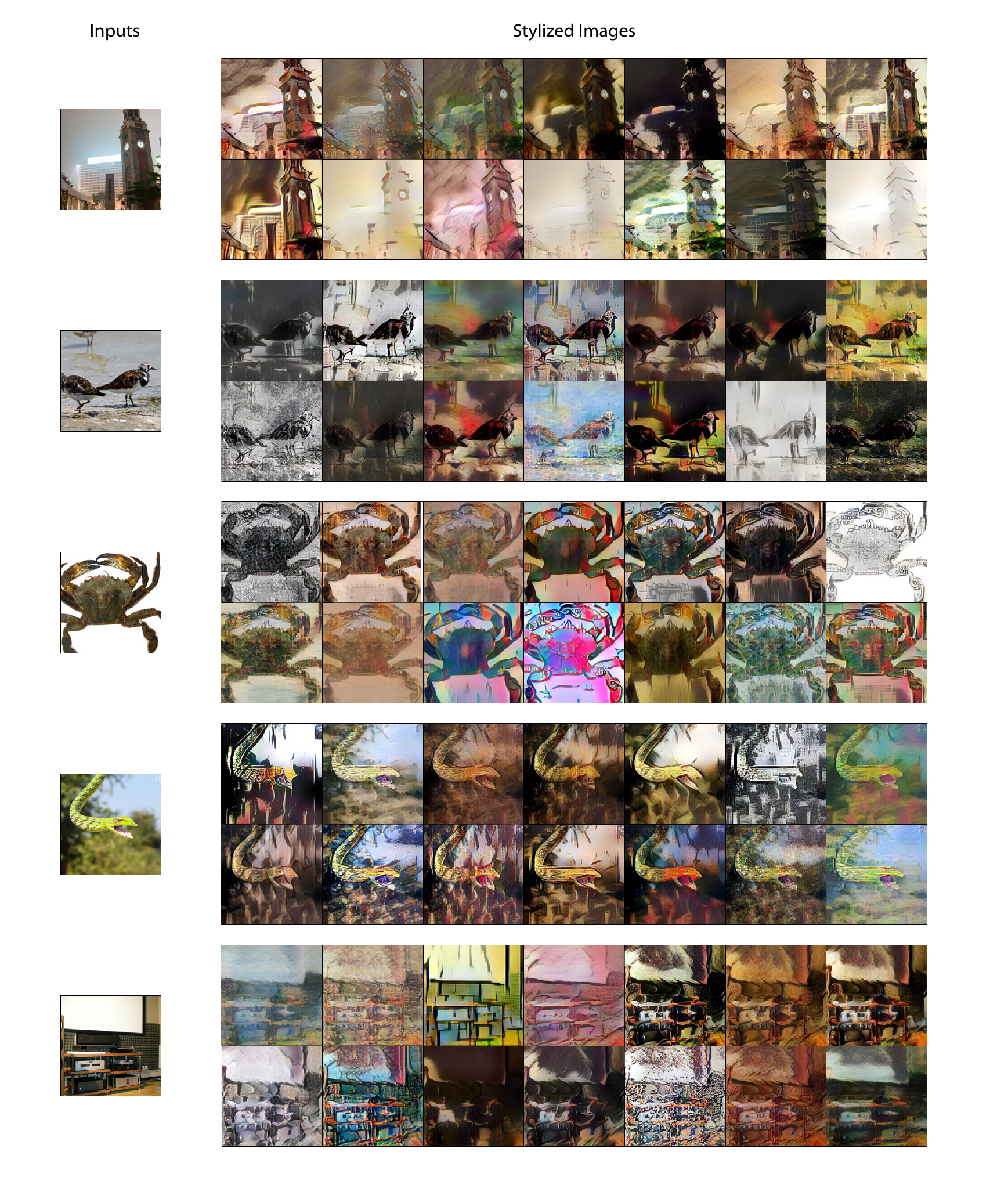}
    \caption{Example images of Stylized ImageNet. We randomly selected 14 different styles for each input image.}
    \label{fig:sin_supp_latter}
\end{figure*}

\begin{figure*}[h]
    \centering
    \includegraphics[width=\textwidth]{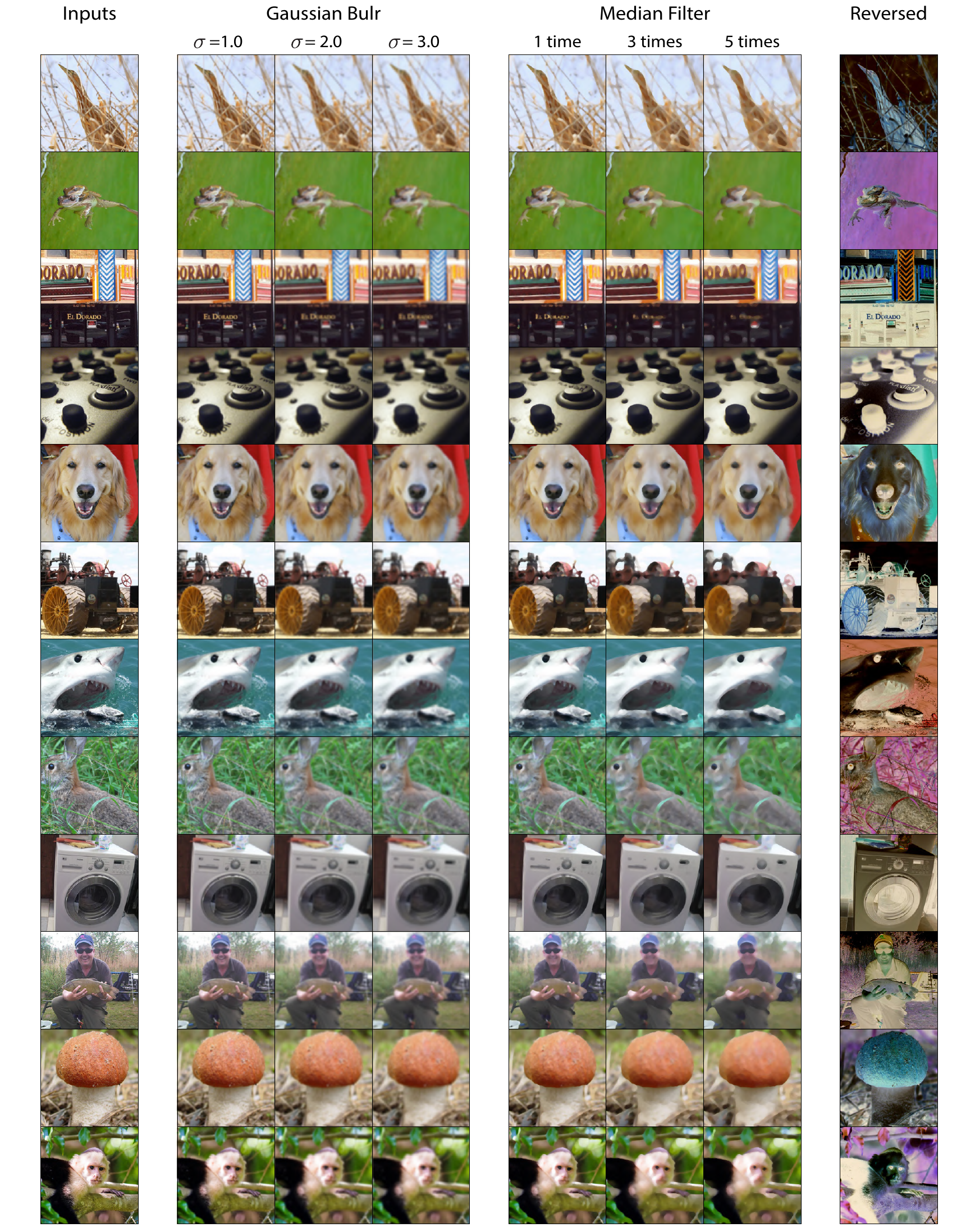}
    \caption{Example images of blurred variant (Gaussian blur and median filter) and color reversed variant of ImageNet.}
    \label{fig:blur_median_supp}
\end{figure*}

Here we provide some example images of variants on ImageNet that are used in our experiments. In Figure~\ref{fig:sin_supp_formar} and Figure~\ref{fig:sin_supp_latter}, we represent some examples of stylized variants of ImageNet that are generated in the same way as Stylized ImageNet~\cite{geirhos2018imagenet}. As we can see, after applying style transfer, local texture cues are no longer highly predictive of the target class, although the global shapes and strong edges tend to be retained.
Next, we provide the sample images that applied with Gaussian blur and median filter in Figure~\ref{fig:blur_median_supp}. While the edges in Gaussian blurred images become blurry equally, the median filter only removes weak edges and noise. Figure~\ref{fig:blur_median_supp} also includes example images of color reversed variant of ImageNet.

\end{document}